\documentclass[10pt,twocolumn]{IEEEtran}

\usepackage{amsmath,graphicx}
\usepackage{algorithm,algorithmicx}
\usepackage{algpseudocode}
\usepackage{subfigure}
\usepackage{color}
\usepackage[]{mcode}
\usepackage{overpic}
\usepackage{pdfcomment,todonotes}

\begin{document}

\title{Weighted Mean Curvature}

\author{Yuanhao Gong and Orcun Goksel
	\thanks{Yuanhao Gong and Orcun Goksel are with Computer-assisted Applications in Medicine, Computer Vision Lab, ETH Zurich, Switzerland}}

\markboth{Gong and Goksel: Weighted Mean Curvature}{Gong and Goksel: Weighted Mean Curvature}
\IEEEaftertitletext{\vspace{-1.5\baselineskip}}

\maketitle
\begin{abstract}
In image processing tasks, spatial priors are essential for robust computations, regularization, algorithmic design and Bayesian inference.
In this paper, we introduce weighted mean curvature (WMC) as a novel image prior and present an efficient computation scheme for its discretization in practical image processing applications. 
We first demonstrate the favorable properties of WMC, such as sampling invariance, scale invariance, and contrast invariance with Gaussian noise model; and we show the relation of WMC to area regularization. 
We further propose an efficient computation scheme for discretized WMC, which is demonstrated herein to process over 33.2\,giga-pixels/second on GPU.
This scheme yields itself to a convolutional neural network representation. Finally,
WMC is evaluated on synthetic and real images, showing its superiority quantitatively to total-variation and mean curvature.
\end{abstract}

\section{Introduction}

Recovering a signal from one or more observations is a fundamental task in image processing, such as in denoising, super-resolution, deconvolution, dehazing, and enhancement. 
The act of generating an observation from the physical space (or an original signal) is called the imaging process, while the model governing this process is the imaging model.
Prior information on this model is a fundamental piece of assumption, which can determine the success or failure of image processing methods.
A typical example is \emph{image denoising}, where the observed data contains measurement errors or noise, which is commonly assumed to follow an expected distribution, given some prior knowledge on the imaging (observation) process.
In \emph{image smoothing}, the goal is to remove undesired details while preserving ``major'' structures, where the structure-detail differentiation is again based on some assumed priors.

Among priors, the well-known Total Variation (TV) assumes that the signal or image to be recovered is a piecewise-constant function~\cite{TV1992}, which has been used successfully in many image processing tasks over the years.
Another potential prior is the assumption that the original image has minimal area~\cite{Graber2015}, which enforces both the gradient and the normal changes to be small, hence resulting in smoother images.  However, this minimal-area assumption is difficult to apply using conventional optimization algorithms. Alternatively, a general gradient distribution prior can be assumed, which imposes the gradient to satisfy certain distributions, rather than being minimized~\cite{gong:gdp}. 
Beyond first-order information, higher-order quantities such as curvature can also be used as a prior. 
Popular choices are Gaussian curvature~\cite{gong2013a} and mean curvature~\cite{MC:1998,LLS,gong:cf,gong:phd}. 
Since these higher-order priors already assume the original signal is higher-order differentiable, the resulting images are enforced to be smooth; consequently, often losing any sharp edge detail. 

For an image processing task, even when the imaging model is the same, the use of different priors may lead to different results. 
For example, assuming TV will lead to piecewise-constant image results; while with area regularization, the final estimation will be close to a piecewise-minimal surface.
Similarly, for Gaussian curvature regularization, the result would be close to a piecewise developable surface~\cite{gong:cf}. 
Accordingly, the prior and applied regularization may affect the results that can be expected from an image processing task in a major way, and a suitable prior is a crucial choice.

Prior choice may also be affected by the available or affordable solution strategy, since imposing some priors require specific numerical solvers for computation. 
For instance, many solvers have been developed for TV for algorithmic efficiency, including the primal-dual method~\cite{chambolle:2011}, split-Bregmann method~\cite{goldstein2009geometric}, and the alternating direction method of multipliers (ADMM). 
These solvers nevertheless cannot be easily extended for other efficacious
priors such as mean curvature or Gaussian curvature~\cite{zhu:2013,gong:cf,shen2003,gong:phd}.
As a result, physically-natural priors representative for many imaging models and with advantageous properties and efficient solution strategies are an ever-existing need.

Among image priors, mean curvature (MC) and its variants are especially interesting, as mean curvature is the gradient of TV term and is further related with the classical mean curvature flow~\cite{chen1991} and the Euler elastic bending energy~\cite{shen2003}.
A special case of Euler elasticity, called Willmore energy, is used in computer graphics for triangular mesh processing~\cite{Crane:2013,jing-yi:2013}. 
Willmore energy is also preferred by most cell membranes~\cite{zimmerberg:2006,velimirovic:2010}, demonstrating its biomechanical relevance. 

In mathematics, MC is the average of all principal curvatures. It is also the average of all curvatures of cross-sectional curves created on that surface when it is cut at a point by all possible planes rotated around the surface normal at that point.
This relates to the diffusion of heat-type equations, and in fluid mechanics, to the equilibrium of spherical droplets.

In this paper, we present a \emph{weighted mean curvature} (WMC) prior that has several advantages properties, including sampling- and scale-invariance, its sparsity on natural images and relation to the gradient of a typical regularization. 
These properties are presented on example images and WMC is compared with traditional priors in numerical experiments on synthetic and real images.
Since WMC does not assume the image smoothness, it is shown to successfully preserve sharp details in the resulting images.
We further present a novel computation scheme to efficiently calculate WMC, including a neural-network implementation for GPU computations. 

\section{Background and Motivation}

Many image processing models can be expressed with the following variational framework:
\begin{equation}
U=\arg\min_U\left\{\, {\cal E}(U(\vec{x}),f(\vec{x}))+ \lambda\,{\cal R}(U(\vec{x})) \,\right\}\,,
\end{equation} 
where ${\cal E}$ is the imaging model, $U$ is the unknown image to be estimated, $f$ is the observed data, $\vec{x}$$\in$$R^n$ is the spatial coordinate, $n$ is the image dimension, $\lambda$$>$$0$ is a parameter (usually related with noise level), and ${\cal R}$ is the regularization term that imposes the assumed prior information. 
The prior term ${\cal R}$ is often independent from the imaging model ${\cal E}$; i.e., the same prior ${\cal R}$ can be used for various imaging models. 

\subsection{Total Variation Regularization}
\begin{figure}[!tb]
	\centering
	\subfigure[the ratio]{
	\includegraphics[width=.48\linewidth]{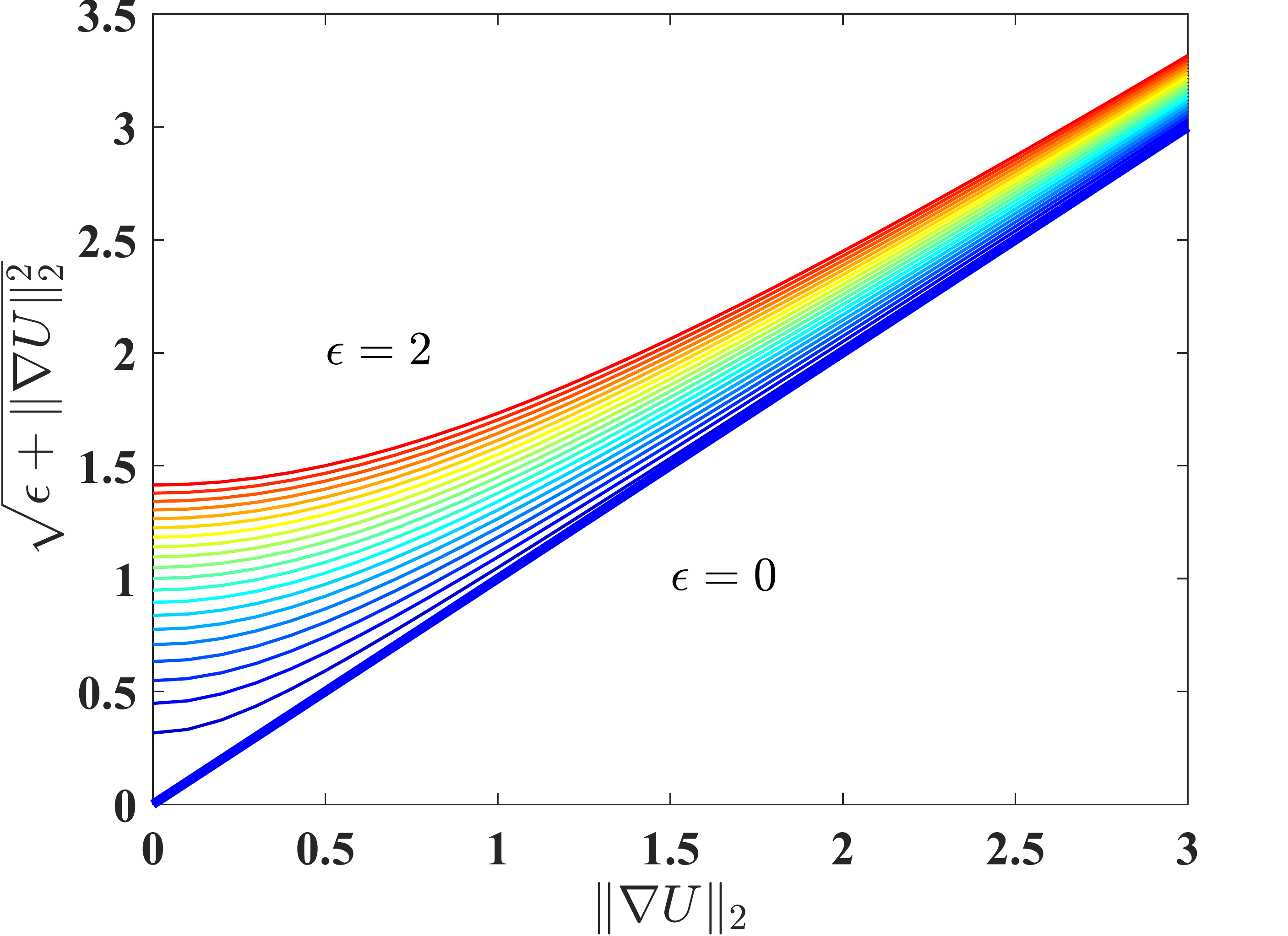}}
\subfigure[the difference]{
	\includegraphics[width=.48\linewidth]{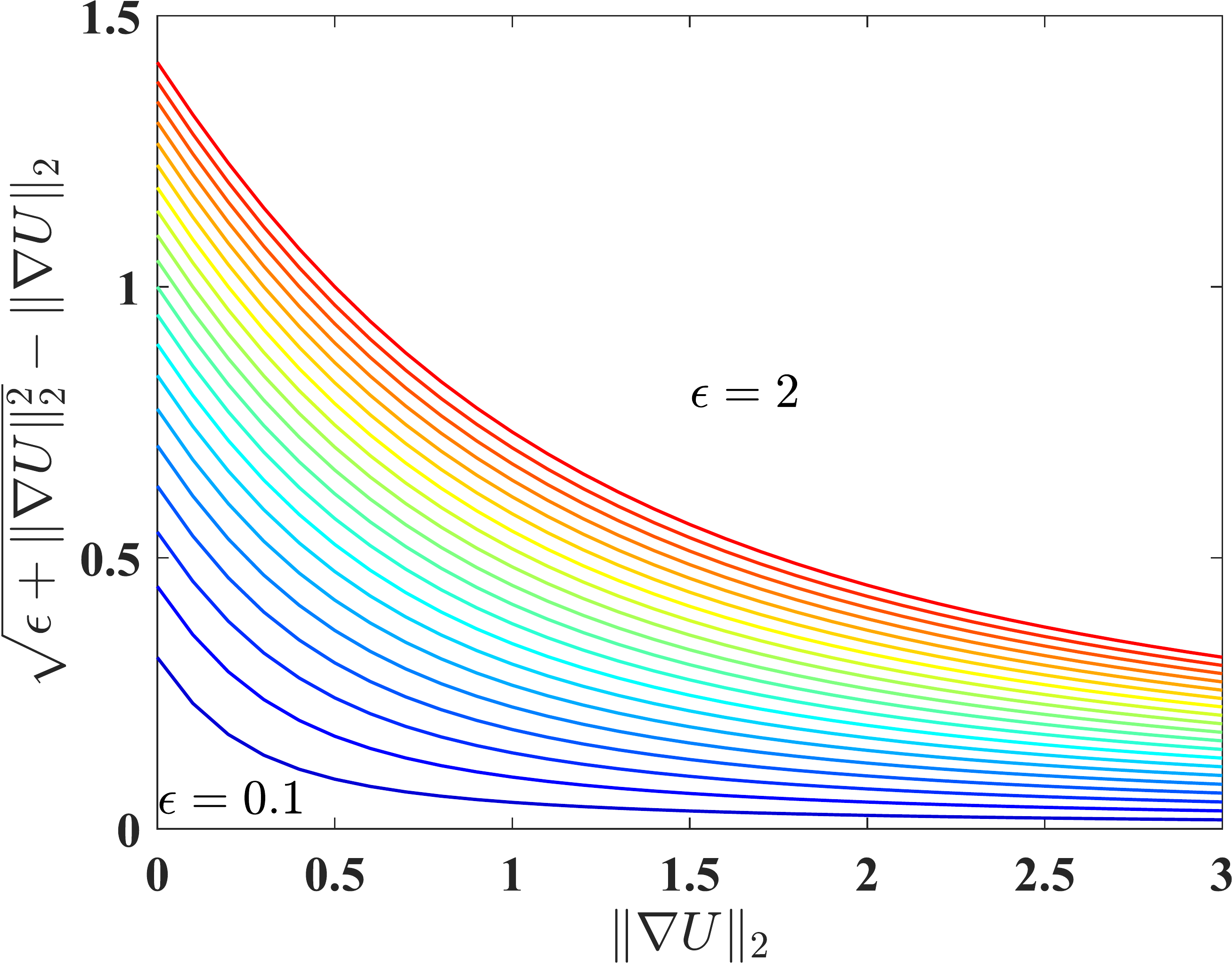}}
	\caption{The role of $\epsilon$ in TV approximation, as $\epsilon$ is increased from 0 to 2 in increments of 0.1 (from blue to red curves).}
	\label{fig:TVeps}
\end{figure}

One of the most popular prior terms is TV regularization 
\begin{equation}
{\cal R}_\mathrm{TV}=\|\nabla U\|_2\,,
\end{equation} 
where $\nabla$ is the gradient operator, and for most further notations, we drop $\vec{x}$ for simplicity.
This regularization has many variants. One of them adopts $\ell_1$ norm to impose anisotropy
\begin{equation}
{\cal R}_\mathrm{TVL1}=\|\nabla U\|_1\,,
\end{equation} 
which prefers horizontal and vertical edges in $U$. 
To improve numerical stability, the following variant is also used:
\begin{equation}
{\cal R}_\mathrm{\sim TV}=\int\sqrt{\epsilon+\|\nabla U(\vec{x})\|_2^2}\ \mathrm{d}\vec{x}\,,
\end{equation} where $\epsilon$$>$$0$ is a small real number. 
Although ${\cal R}_\mathrm{\sim TV}$ numerically approximates ${\cal R}_\mathrm{TV}$, these two terms are fundamentally different. 
This is demonstrated in Fig.~\ref{fig:TVeps} as $\epsilon$ is varied between 0 and 2 in increments of 0.1.
Although, for small $\epsilon$, this approximation error may be small at any given location $\vec{x}$, its integration on the whole imaging domain may add up to a substantial value, especially if the image area is large. 
This highlights the effect of $\epsilon$ beyond its numerical reasons.
 
For $\epsilon=1$, the above becomes the area regularization
\begin{equation} \label{eq:areaReg}
{\cal R}_{\mathrm{area}}=\int\sqrt{1+\|\nabla U(\vec{x})\|_2^2}\ \mathrm{d}\vec{x}\,,
\end{equation}
which imposes that $U$ has a minimum area.
Thus, this requires not only the gradient to be smooth, but also the image normal to be smooth.
As a result, common staircase artifacts from TV can be significantly reduced using ${\cal R}_{\mathrm{area}}$~\cite{Graber2015}. 

Minimizing TV usually requires the gradient of ${\cal R}$ to be computed with respect to $U$~\cite{TV1992}, i.e.
\begin{equation}
\label{eq:tvgrad}
	-\frac{\partial {\cal R}_{TV}}{\partial U}=\frac{1}{2}\nabla\cdot\frac{\nabla U}{\|\nabla U\|_2}\,,
\end{equation} where $\nabla\cdot$ is the divergence operator. The right hand side of Eq.\,(\ref{eq:tvgrad}) is indeed the mean curvature of the isocontours of $U$. 
Note that this computation as such would lead to numerical issues as $\|\nabla U\|$ vanishes. Note that the above definition of mean curvature for contours is different from that for surfaces or graphs, the latter of which was studied in earlier works~\cite{gong:cf,gong:Bernstein}. 
In this paper, we focus on the mean curvature definition commonly used in image community as in Eq.~\ref{eq:tvgrad}, and extend this to address its shortcomings with our proposed definition of \emph{weighted mean curvature}.

\subsection{Mean Curvature in Image Processing}

Mean curvature (MC) has been used extensively in image processing problems~\cite{MC:1998,huisken2001,LLS,gong:cf,gong:phd}, such as denoising~\cite{Zhu2007,zhu:2013}, registration~\cite{zhang_chen_chen_yu_2017}, reconstruction~\cite{RenQL15}, and decomposition~\cite{el20172d}. 
For an image $U(\vec{x})$, the mean curvature of $U$ at $\vec{x}$ is 
\begin{equation}
\label{eq:2d}
H(U)=\frac{1}{n}\nabla\cdot\frac{\nabla U}{\|\nabla U\|_2},
\end{equation} where $\nabla$ and $\nabla\cdot$ are gradient and divergence operators, respectively. 
For $n$$=$$2$, MC is equivalent to the right-hand side of Eq.\,(\ref{eq:tvgrad}) and we have
\begin{equation}
\label{eq:2dlong}
\begin{split}
&H=\frac{U_x^2U_{yy}-2U_xU_yU_{xy} +U_y^2U_{xx}}{2(U_x^2+U_y^2)^{\frac{3}{2}}}\\
&=\frac{U_{yy}+U_{xx}}{2(U_x^2+U_y^2)^{\frac{1}{2}}}-\frac{U_y^2U_{yy}+2U_xU_yU_{xy} +U_x^2U_{xx}}{2(U_x^2+U_y^2)^{\frac{3}{2}}}\,.
\end{split}
\end{equation} 
This equation links MC with the Laplace operator and the diffusion along normal direction.

Mean curvature is independent of image contrast because $H(\alpha U)$$=$$H(U)$ for any scalar $\alpha$$\neq$$0$. 
Thanks to this property, MC can provide a uniform regularization for images that contain objects of different contrast and thus is advantageous as image prior for regularization.

Mean curvature regularization is defined as
\begin{equation}
	{\cal R}_H(U)=\int |H(U)|^q\,\mathrm{d}\vec{x}\,,
\end{equation} 
where $q$$>$$0$ is a scalar parameter; usually set to 1 or 2. 
If ${\cal R}_H(U)$$=$$0$, the corresponding $U$ is then a piecewise minimal surface (i.e., $\forall \vec{x}, H(U)$$=$$0$). 

Compared to TV regularization, ${\cal R}_H$ leads to better results for image denoising in practice~\cite{Zhu2007,meanZhu}, which has been explained theoretically from a geometry point of view in~\cite{gong:cf,gong:phd,gong2009symmetry} and a function analysis point of view in~\cite{gong:Bernstein,gong2013a}.   

\subsection{Challenges with Mean Curvature}
Despite its attractive features, application of MC has several difficulties in practice. 
First, the MC depends on scale and its use as regularization, i.e.\, ${\cal R}_H$, depends on sampling rate (described further in Section~\ref{sec:sample}). 
Thus, MC is not only affected by the geometry itself, but also by the sampling method and scale space, posing challenges as undesired side-effects in practical applications.

Second, minimizing the mean curvature regularization ${\cal R}_H$ is relatively challenging as it leads to a fourth-order partial differential equation~\cite{meanZhu}. 
Although several methods have been proposed, such as the multi grid method~\cite{Carlos:2010}, augmented Lagrange method~\cite{meanZhu,RenQL15}, and the fixed point method~\cite{yang:2014}, to substantially reduced computations, their application on larger images is still far from practical, given realistic amount of computational resources. 

Third, the discretization of Eq.\,(\ref{eq:2dlong}) requires the first and second order derivatives to be approximated, for instance, by finite differencing, which is susceptible to noise and can be highly unstable. 
Further numerical issues arise for vanishing $\|\nabla U\|$ that appears in the denominator.

Finally, $U$ needs to be assumed as smooth when numerically calculating its second order derivative. Although the mean curvature filter was proposed recently to relax this constraint via implicit minimization without computing the high-order derivatives~\cite{gong:cf,gong:Bernstein}, the above computational challenges still persist. 

\subsection{Our Motivation and Contribution}
To overcome the above challenges, we propose herein \emph{weighted mean curvature} (WMC) to be used instead of mean curvature.
WMC is fundamentally different from MC, with its following advantageous properties detailed further in the next section:
\begin{itemize}
	\item {\bf sampling-} and {\bf scale-invariance,} crucial for images that contain objects with different scales and/or of different sampling rates.
	\item{\bf sparsity on natural images:} As is later shown statistically on natural images, WMC is sparser than gradient, and thus would be preferred as a regularizer.
	\item {\bf gradient of area regularization} can be numerically approximated using WMC, significantly simplifying the corresponding optimization procedure.
	\item{\bf fast computation scheme:} For applying our weighted mean curvature on discrete images, we further propose herein a fast discrete computation scheme that can approximate WMC numerically.  This has further advantages, including: (i) it does not require the signal to be second-order differentiable, hence better preserving edges; (ii) it is a very fast discrete operation, making WMC practical for most (even computatinally-demanding) image processing tasks; and (iii) it is numerical stable when $\|\nabla U\|_2$ vanishes and thus avoids any numerical issues, e.g.\ in contrast to those encountered with TV in Eq.(\ref{eq:tvgrad}) and MC in Eq.(\ref{eq:2d}).  
\end{itemize}

\section{Weighted Mean Curvature}
\begin{figure}
	\centering
	\includegraphics[width=\linewidth]{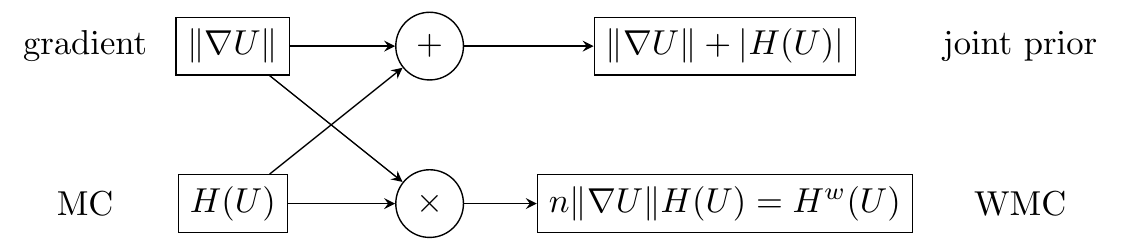}
	\caption{Illustration of relationship between $\nabla U$, $H(U)$, and $H^w(U)$.}
	\label{fig:role}
\end{figure}

We define WMC as
\begin{equation}
\label{eq:wmc}
	H^w(U)=n\|\nabla U\|_2\,H(U)=\|\nabla U\|_2\left(\nabla\cdot\frac{\nabla U}{\|\nabla U\|_2}\right)\,.
\end{equation} For 2D images $n=2$, this equation becomes
\begin{equation}
\label{eq:rhs}
H^w(U)=\underbrace{~\Delta U~}_{\text{isotropic diffusion}} - \underbrace{\frac{U_y^2U_{yy}+2U_xU_yU_{xy} +U_x^2U_{xx}}{U_x^2+U_y^2}}_{\text{diffusion along normal direction}}\,,
\end{equation} where $\Delta$ is the isotropic Laplace operator. Although this term still has the problem when $\|\nabla U\|_2=0$, its discrete computation can avoid such issue as shown later in Section~\ref{sec:fast}. 
This WMC definition can be interpreted either as {\bf gradient \emph{weighted by} mean curvature} (weighted Total-Variation) or as {\bf mean curvature \emph{weighted by} gradient}. 
The relationship of WMC to gradient and MC is illustrated graphically in Fig.~\ref{fig:role}, in order to emphasize that the multiplication of the terms makes WMC fundamentally different from both gradient and mean curvature.

For WMC it is apparent that when \emph{either} contributing term vanishes, their multiplication WMC will also go to zero. The small value of TV is a sufficient condition for the small value of WMC. Therefore, WMC is sparser i.e.\ is statistically more likely to be zero compared to the TV.
We present a statistical comparison of these given natural images in Section~\ref{sec:stat}. 

The above implies that WMC regularization is a superset of both ${\cal R}_{H}$ and ${\cal R}_{TV}$. 
WMC regularization term can be defined as
\begin{equation}
\label{eq:wmcr}
{\cal R}_{H^w}(U)=\int\bigl|H^w(U)\bigr|^q\mathrm{d}\vec{x}=\int\bigl|H(U)\,\|\nabla U\|_2\bigr|^q\mathrm{d}\vec{x}\,,
\end{equation} where $q$$>$$0$ is a scalar parameter defining the norm. 
We use $q$$=$$1$ unless explicitly stated. 
Minimizing our WMC regularization requires either gradient or mean-curvature to be small. In other words, it will allow one of them to be large while the other one is close to zero. This behavior allows ${\cal R}_{H^w}$ to \emph{automatically} choose gradient or MC or both to be minimized, leading to less artifacts compared to MC regularization ${\cal R}_{H}$ and TV regularization ${\cal R}_{TV}$.

In this paper, we introduce and study WMC as it can be used as both a regularization as well as the gradient of some regularization as shown in the following section.
To that end, efficient numerical computation of WMC is essential, which is later addressed in Section~\ref{sec:fast}.
We also show below the sampling and scale invariance properties of WMC regularization.
The gradient of WMC regularization, however, is not straight-forward to calculate, which is left for future work for potential application scenarios that may necessitate that.
Nevertheless, we show several other practical applications in our results where WMC itself can be successfully computed and used.

\subsection{Sampling- and Scale-Invariance}
\label{sec:sample}

Note that while the MC regularizer in Eq.(9) is an integration of individual curvatures, the WMC regularizer we proposed is normalized (weighted) by divergence magnitude, cf.\ Eq.(12).
Therefore, the spatial sampling effect is taken into account inherently by WMC.
Similarly, the effect of scale is also incorporated through this weighting in WMC, which makes it a physically meaningful quantity.  We further demonstrate this on a toy example below.

\begin{figure}[!tb]
	\centering
	\includegraphics[width=.7\linewidth]{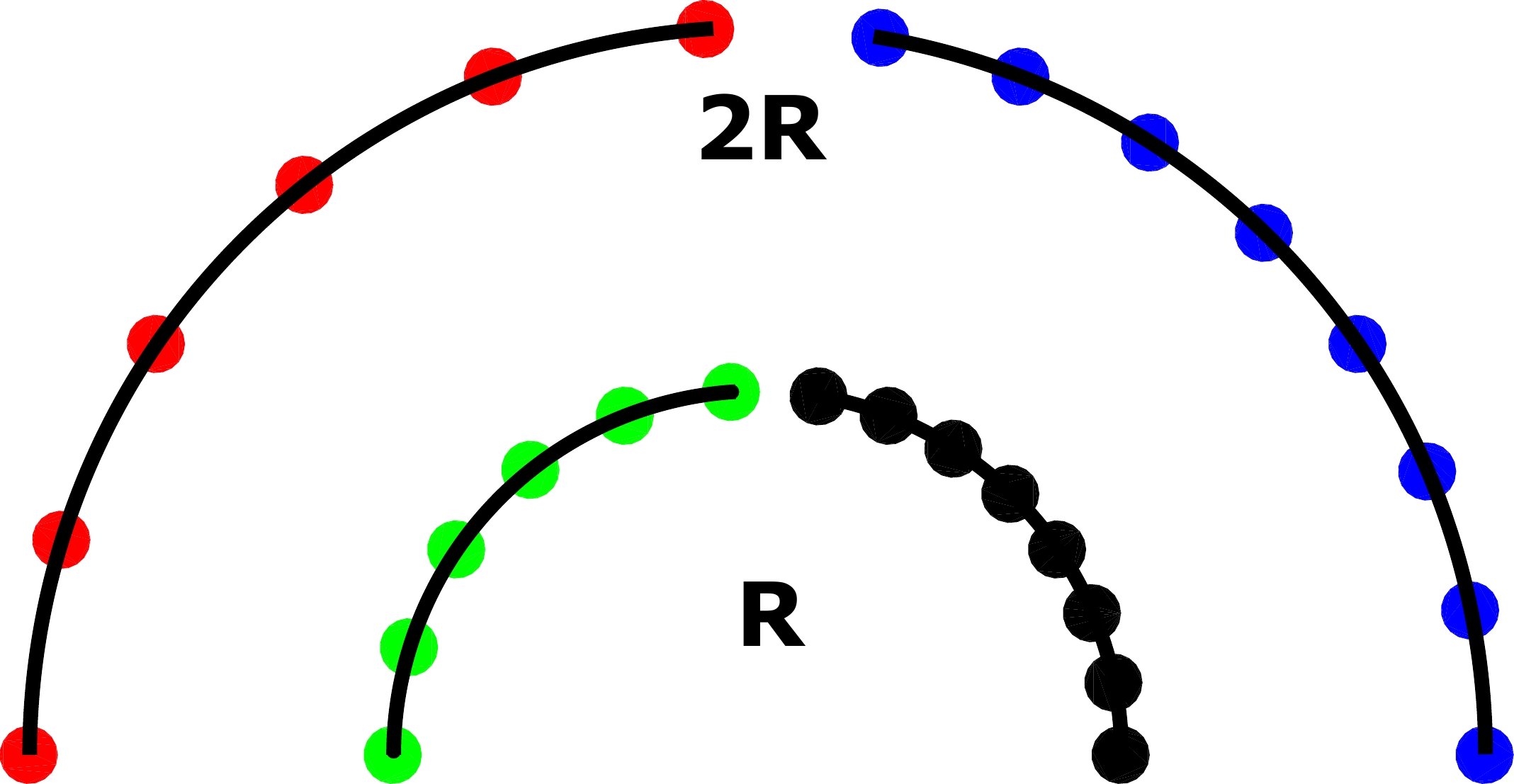}
	\caption{Different sampling rates on half circles with radii $R$ and $2R$.}
	\label{fig:half_circles}
\end{figure}

Consider arcs with different sampling rates at two scales in Fig.~\ref{fig:half_circles}, where gradients and curvatures of the arcs are computed at the given samples along these segments. 
For the inner (green and black) arcs of $R$ radii, curvature $H$ at the samples on the left and right segments are both equal, i.e.\ $\frac{1}{R}$. This then yields their integration ${\cal R}_H$ to be different as the number of integration points (samples) differ, i.e.\ $\frac{6}{R}$ vs. $\frac{8}{R}$ for the given examples, despite the fact that they present the same underlying geometry and continuum. In contrast, ${\cal R}_{H^w}$ yields the same value
\begin{equation}
	{\cal R}_{H^w}\approx\sum_{i=1}^{6}\frac{1}{R}\frac{\pi R}{4\times 6}=\sum_{i=1}^{8}\frac{1}{R}\frac{\pi R}{4\times 8}\,.
\end{equation}
demonstrating the sampling-invariance of WMC regularization ${\cal R}_{H^w}$.

Consider the scaled versions of these arcs, i.e.\ the outer segments with $2R$ radii. 
$H$ would be reduced in both cases to $\frac{1}{2R}$ and thus each ${\cal R}_H$ also been halved.
In contrast, $H^w$ does not change as $\|\nabla U\|_2$ increases proportionally, demonstrating the scale-invariance of both $H^w$ and ${\cal R}_{H^w}$. 

The above presented shortcomings of mean-curvature is conventionally treated via heuristic-methods in image processing community, which is handled intrinsically by our WMC.

\subsection{Contrast Invariance}
Below we first analyze the classical TV model, showing MC as not contrast invariant.
Consider a generic imaging model $AU=f$ where $f$ is observed data, $A$ is the imaging matrix, and $U$ the image.
This commonly has the solution strategy
\begin{equation}
\label{eq:tvopt}
	U^{\star}=\arg\min_U\left\{\frac{1}{2}\|AU-f\|_2^2+\lambda\|\nabla U\|_2\right\}\,.
\end{equation} 
The optimal solution $U^{\star}$ must satisfy Euler-Lagrange equation
\begin{eqnarray}
	A^T(AU^{\star}-f)-\lambda\nabla\cdot\frac{\nabla U^\star}{\|\nabla U^\star\|_2}&=&0\\
\label{eq:eu-h}A^T(AU^{\star}-f)-n\lambda H(U^\star)&=&0\,.
\end{eqnarray}

Note that, since $H(\alpha U)=H(U)$, a scaled solution $\alpha U^\star$ does not satisfy the above for scaled data $\alpha f$, for any $\alpha\neq 0$.
Therefore, the above is not contrast invariant, i.e.
\begin{equation}\label{eq:eu-ah}
A^T(A\alpha U^{\star}-\alpha f)-n\lambda H(\alpha U^{\star}) \neq 0\,.
\end{equation}
Nevertheless, if MC in the above equation is replaced with WMC, contrast invariance can be shown as
\begin{equation}
\label{eq:scale}
A^T(A\alpha U^{\star}-\alpha f)-\lambda H^w(\alpha U^{\star})= 0\,.
\end{equation} 
Note that this is the Euler-Lagrange equation of \emph{some} optimization form, i.e.
\begin{equation}
U^{\star}=\arg\min_U\left\{\frac{1}{2}\|AU-f\|_2^2+\lambda{\cal R}(U)\right\}\,.
\end{equation} 
given \emph{some} regularization term $\cal R$.
The form of such unknown regularization could be nontrivial; nevertheless for any solution strategy, its gradient is the main concern -- which is defined above as WMC.
In the following subsection, this unknown regularization term is shown to be a variant of area regularization and therefore WMC to be an approximation to the gradient of area regularization.

\subsection{Approximate the Gradient of Area Regularization}
\label{sec:grad}

Noticing the similarity between $H^w$ in Eq.~\ref{eq:rhs} and the gradient of area regularization in Eq.~\ref{eq:areaReg}, i.e.
\begin{align}
\label{eq:area_reg}
	-\frac{\partial {\cal R}_{\mathrm{area}}}{\partial U}&=\sqrt{1+|\nabla U|^2}\nabla\cdot\frac{\nabla U}{1+|\nabla U|^2}\\
	\label{eq:area_reg1}
	&=\Delta U - \frac{U_y^2U_{yy}+2U_xU_yU_{xy} +U_x^2U_{xx}}{1+U_x^2+U_y^2}\\
	\label{eq:area_reg2}
	&\approx  \Delta U -\frac{U_y^2U_{yy}+2U_xU_yU_{xy} +U_x^2U_{xx}}{U_x^2+U_y^2}\\
	&=H^w\,,
\end{align} 
we can consequently treat WMC as a numerical approximation to the gradient of the area regularization.
This numerical approximation reduces computations significantly (thanks to the fast computation scheme presented in Section~\ref{sec:fast}), which will enable many image processing tasks to apply area regularization in practical settings.

\begin{figure}
	\centering
	\includegraphics[width=.7\linewidth]{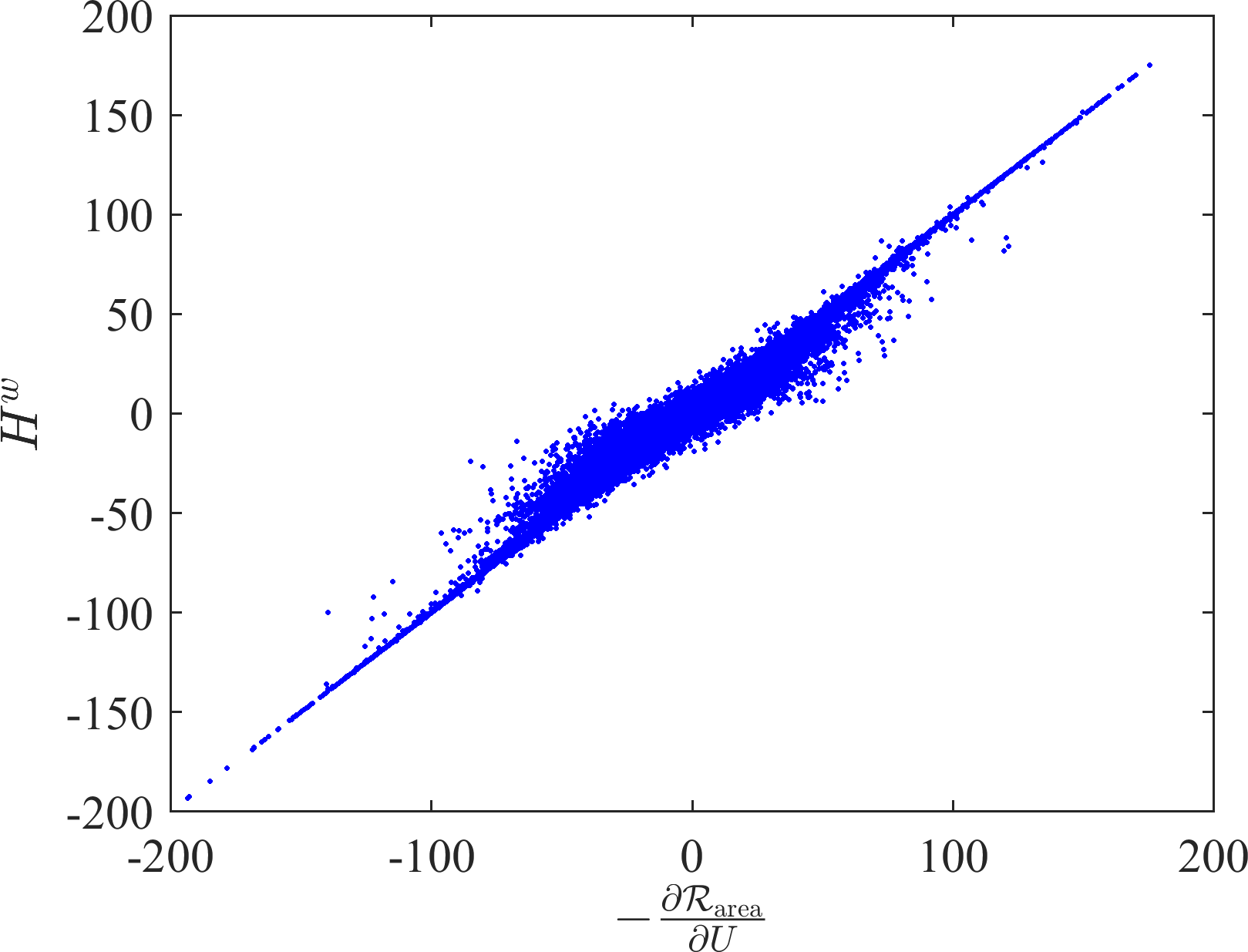}
	\caption{The relationship between $-\frac{\partial {\cal R}_{\mathrm{area}}}{\partial U}$ and $H^w$ on 500 natural images from BSDS500 dataset.}
	\label{fig:errorHW}
\end{figure}

Although the approximation in~(\ref{eq:area_reg2}) is less accurate when the gradient magnitude $|\nabla U|$ vanishes for constant regions of the image, in this case (\ref{eq:area_reg}) becomes the Laplace operator, while $H^w$ per definition in (\ref{eq:wmc}) also becomes the Laplace operator due to imposed Neumann boundary conditions; making these two quantities similar.
We analyze WMC approximation to the gradient of the area regularization from two aspects: First, from the function analysis point of view, the approximation error is reduced when the gradient norm gets larger, as can be seen from the equations (\ref{eq:area_reg1} and \ref{eq:area_reg2}). 
Second, the relationship between $-\frac{\partial {\cal R}_{\mathrm{area}}}{\partial U}$ and $H^w$ can also be shown statistically, e.g. in Fig.~\ref{fig:errorHW} as computed on 500 natural images from BSDS500 dataset~\cite{Arbelaez2011}. 

Based on the above approximation, the following model becomes \emph{contrast-robust} (although not invariant due to the approximation)
\begin{equation}
U^\star=\arg\min_U\left\{\frac{1}{2}\|AU-f\|_2^2+\lambda{\cal R}_{\mathrm{area}}\right\}\,.
\end{equation}
Practical numerical applications of this model are shown in Section~\ref{sec:compare}.

\subsection{Statistics of WMC on Natural Images}
\label{sec:stat}

\begin{figure}[b]
	\centering
	\subfigure[distributions in log scale]{\includegraphics[width=.47\linewidth]{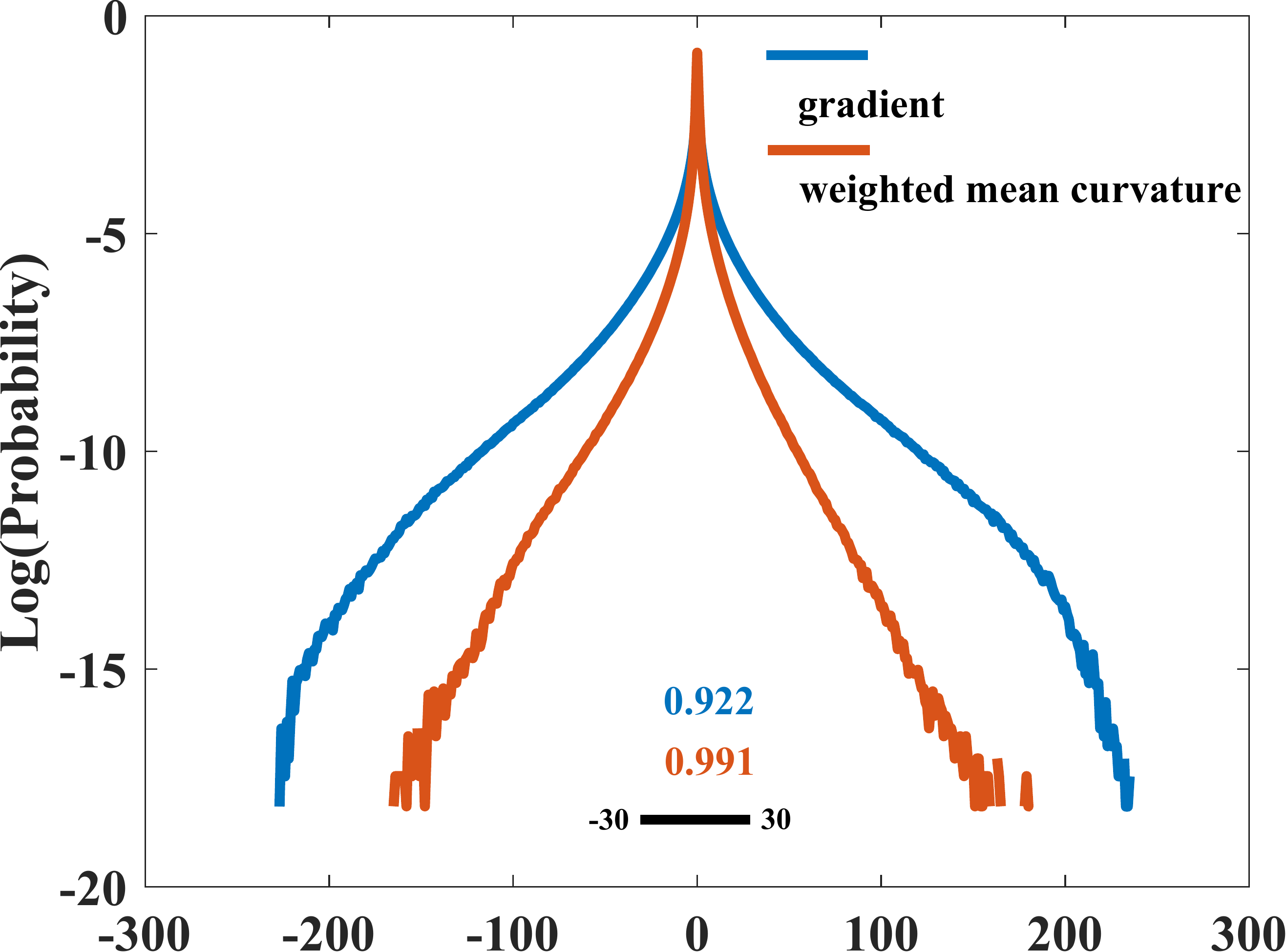}}
	\subfigure[cumulative distribution]{\includegraphics[width=.47\linewidth]{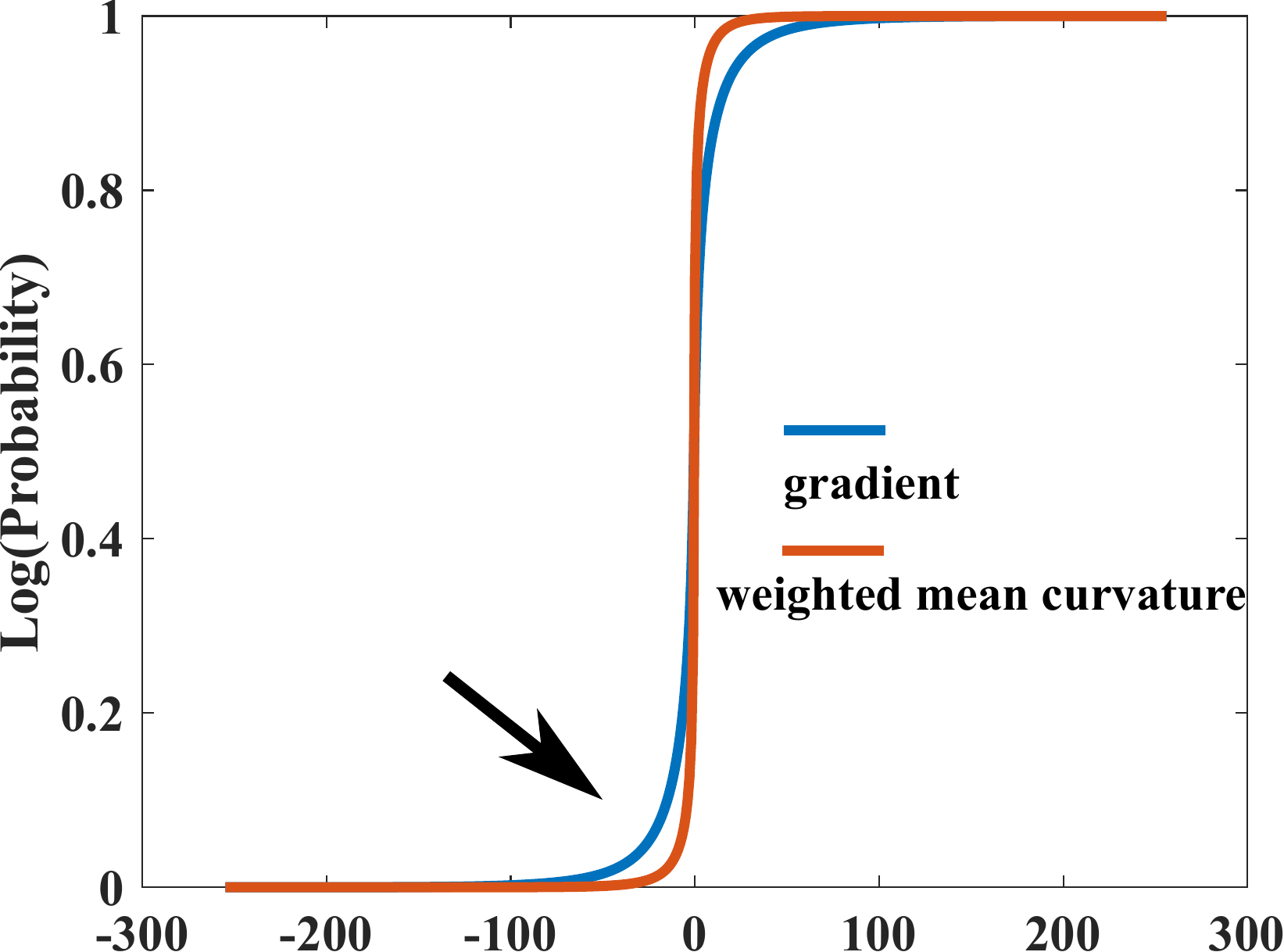}}
	\subfigure[distribution modeling]{\includegraphics[width=.45\linewidth]{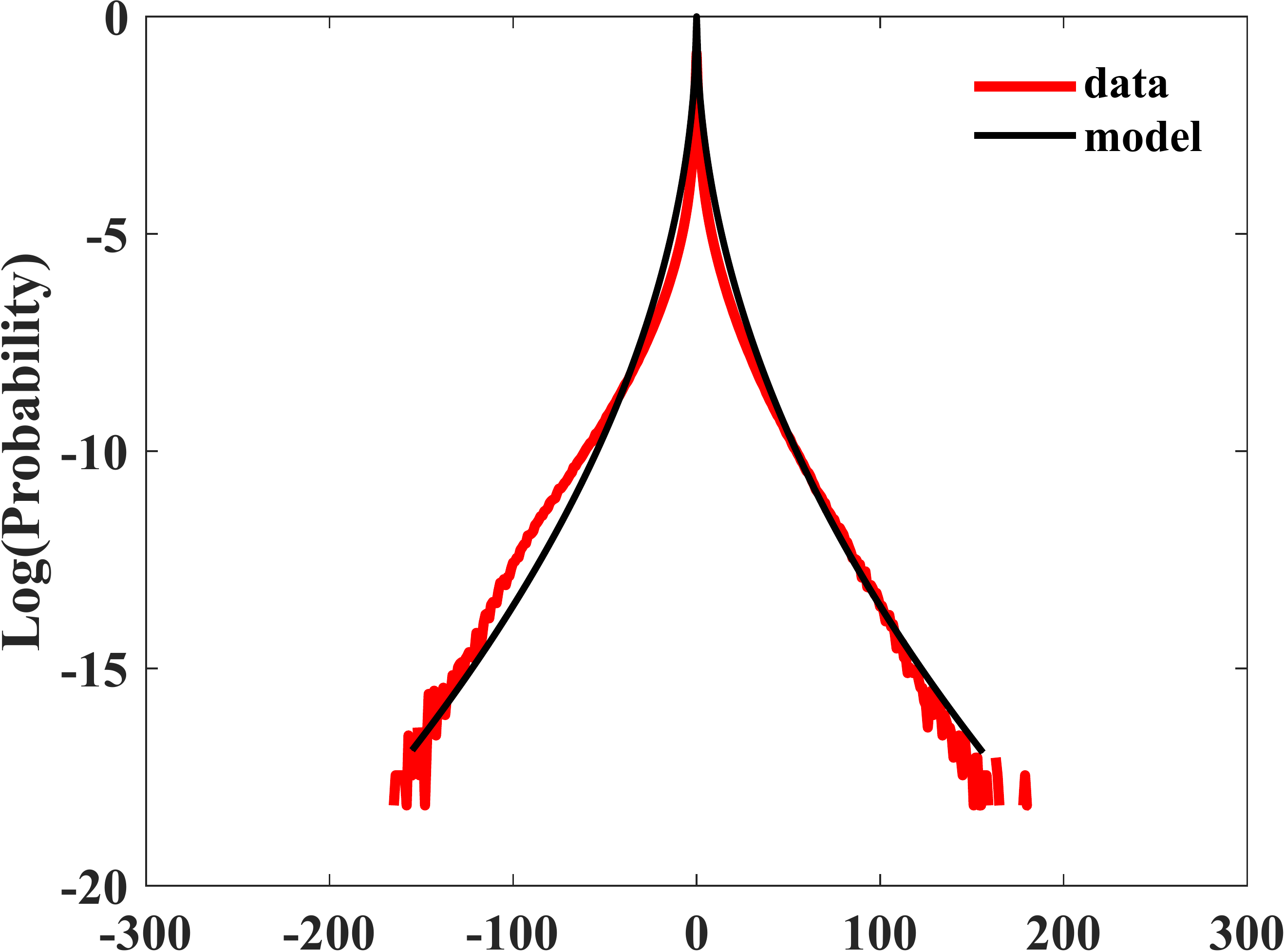}}
	\caption{Sparsity demonstration by (a) log-probability and (b) cumulative distributions of WMC and gradient, also (c) using a fitted model.}
	\label{fig:stat}
\end{figure}

It is well-known that the gradient in natural images exhibits a heavy tail distribution. The mean curvature MC also presents a similar distribution~\cite{gong:phd}. Here, we show that WMC $H^w$ satisfies a similar but even sparser distribution. 
We used 500 natural images from BSDS500 dataset. 
We computed (axis-aligned) gradients and WMC in these images, and in Fig.~\ref{fig:stat}(a) we show the average gradient and average WMC distributions in log scale. 
It shows that WMC is sparser than gradient in this dataset. More specifically, let $p(H^w)$ and $p(\nabla U)$ denote the probability of WMC and the probability of gradient, respectively. Then, it can be seen that for any given real positive threshold $v$, the following holds: $p(|H^w|\leq\mathrm{v})>p(|\nabla U|\leq\mathrm{v})$. 
For example, for a threshold of $v=30$, the probability of WMC is $p(|H^w|\leq 30)=0.991$ while the probability of gradient is $p(|\nabla U|\leq 30)=0.922$, as shown in Fig.~\ref{fig:stat}(a). The sparsity can also be seen from the cumulative distributions in Fig.~\ref{fig:stat}(b), where WMC raises tighter around zero, meaning it is more sparse (i.e. considering absolute values below any given threshold as practically zeros, then WMC will have more of those values).
This fact also corroborates with the theoretical discussion earlier in the paragraph below Fig.~\ref{fig:role} on observing zeros statistically.

Furthermore, the average WMC distribution can be modeled as
\begin{equation}
	-\log (p(H^w))\approx \frac{11}{8}|H^w|^{\frac{1}{2}}\,,
\end{equation} 
as shown the black line in Fig.~\ref{fig:stat}(c), where $p$ represents the probability.
The power of $|H^w|$ term in this model fitting suggests that the parameter $q$ in Eq.~\ref{eq:wmcr} should be close to $\frac{1}{2}$ when using ${\cal R}_{H^w}$ as regularization for natural images. Although $q=\frac{1}{2}$ is better from a modeling point of view, $q$$=$$1$ is preferred from optimization point of view, since $|\cdot|$ is easier than $\sqrt{\cdot}$ to minimize.

\subsection{Connection with Mean Curvature Flow}
Conventional mean curvature flow minimizes the total surface area by evolving a surface according to
\begin{equation}
\frac{\partial U(\vec{x},t)}{\partial t}=H(U)\|\nabla U\|_2\,.
\end{equation}
Note that the above is indeed a normalized form of WMC, i.e.
\begin{equation}
H(U)\|\nabla U\|_2 = \frac{1}{n}H^w(U)\,.
\end{equation}

With a discrete time step size of $\delta$, we then have the iterations
\begin{equation} \label{eq:MCF}
U^{t+1}=U^t+\delta  H^w(U^t)\,.
\end{equation}
This equation can be seen as minimizing the area regularization alone without a data fitting term. 
The bottleneck in performing this iteration is the evaluation of WMC, which conventionally requires an approximation of the first and second derivatives of $U$, which further suffers from numerical issues near $\|\nabla U\|_2=0$.
Nevertheless, this is overcome by the novel computation scheme proposed herein to efficiently approximate WMC on images with a discrete Laplace operator.

\section{Fast Discrete Computation Scheme for WMC}
\label{sec:fast}
We first show the connection between WMC and Laplace operator, and next analyze discrete Laplace kernels and normal directions.
We then present the combination of a regression kernel with discrete normal directions as a fast computation scheme.

\subsection{Discretization of Weighted Mean Curvature}
Given the WMC expression in Eq.(\ref{eq:rhs}), the first term can be approximated by a Laplace operator and the second term is the diffusion along the normal direction.
\subsubsection{Discrete Isotropic Laplace Operator}
\begin{figure}
	\centering
	\subfigure[$k_1$]{\includegraphics[width=.24\linewidth]{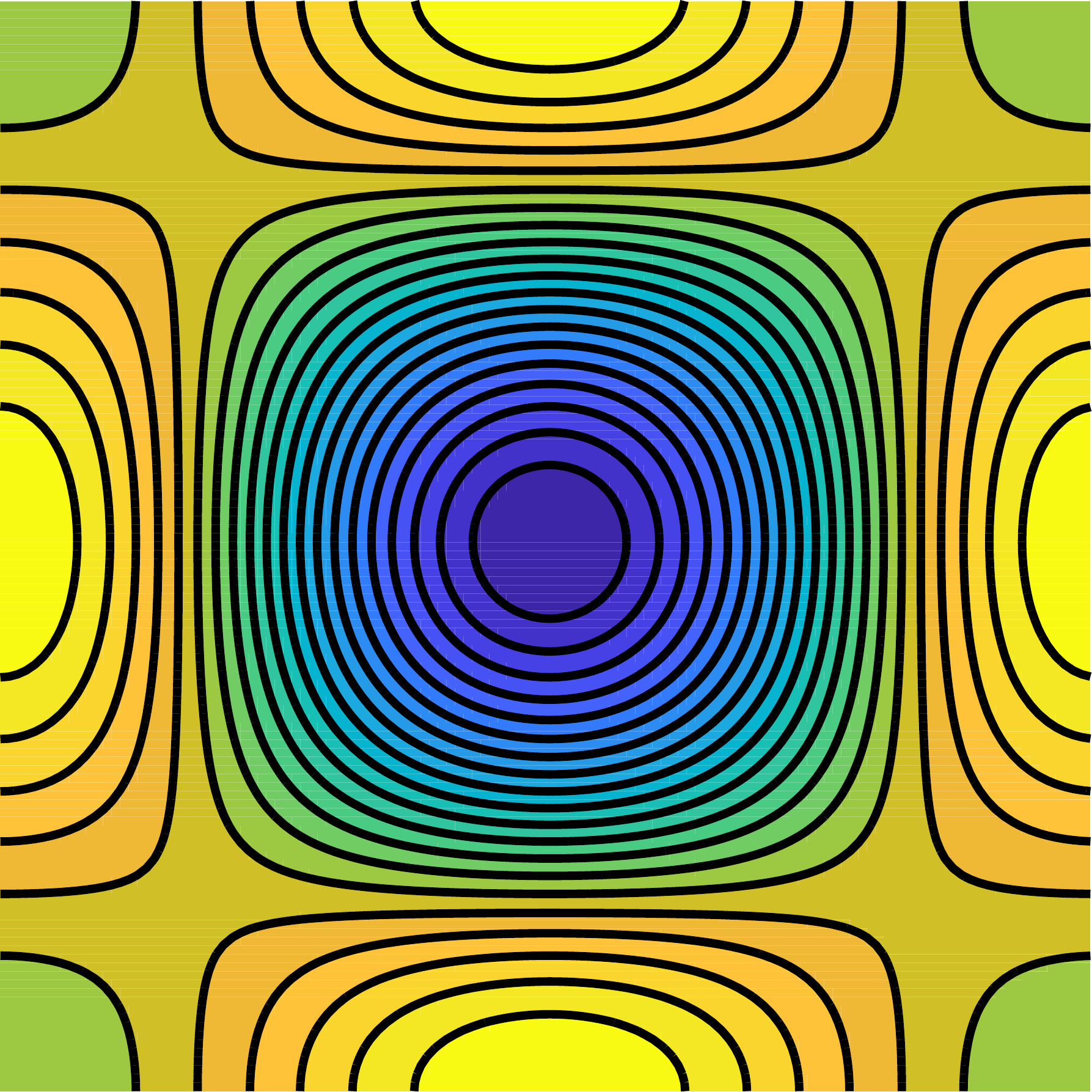}\label{fig:h1}}
	\subfigure[$k_2$]{\includegraphics[width=.24\linewidth]{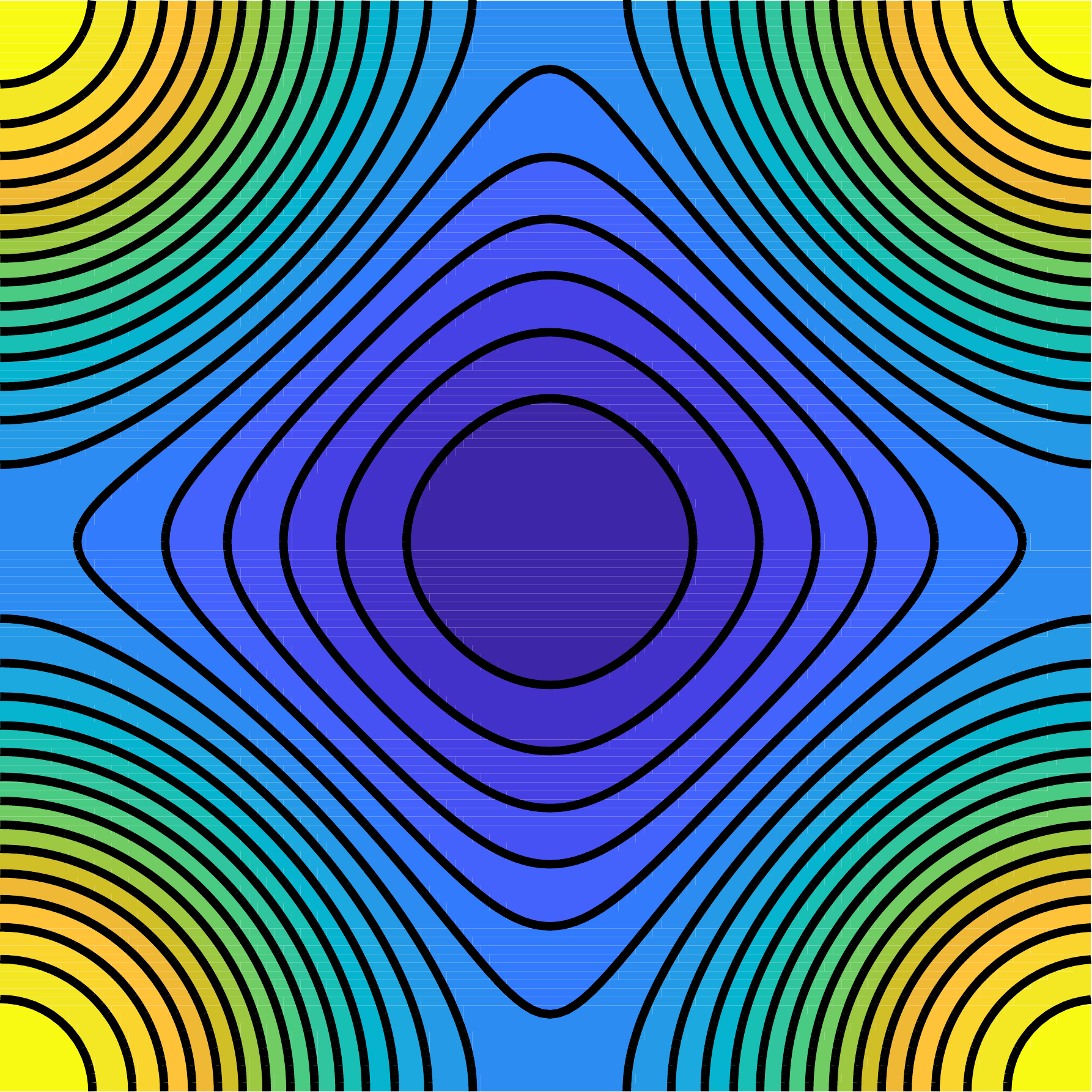}\label{fig:h2}}
	\subfigure[$k_3$]{\includegraphics[width=.24\linewidth]{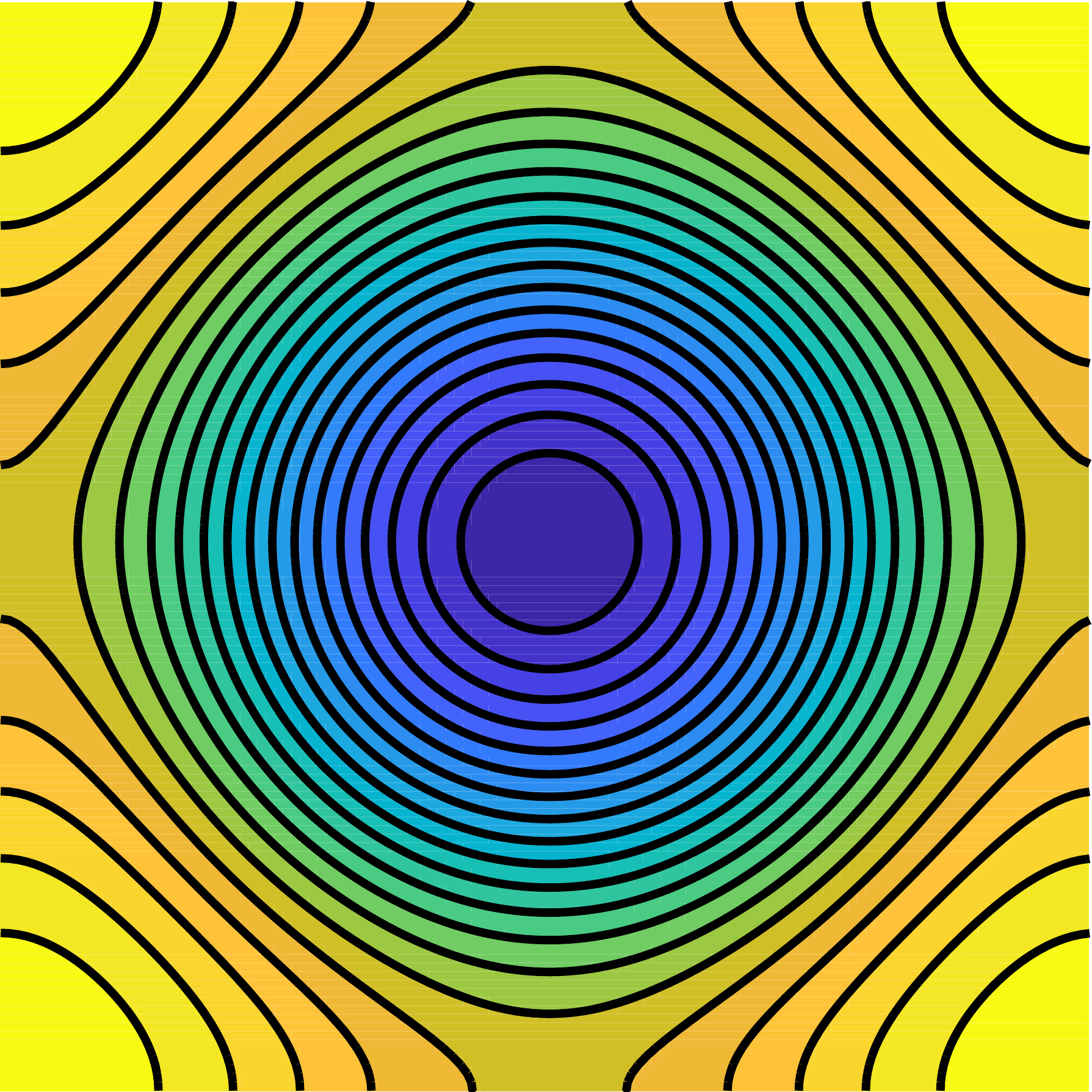}\label{fig:h3}}
	\subfigure[$k_4$]{\includegraphics[width=.24\linewidth]{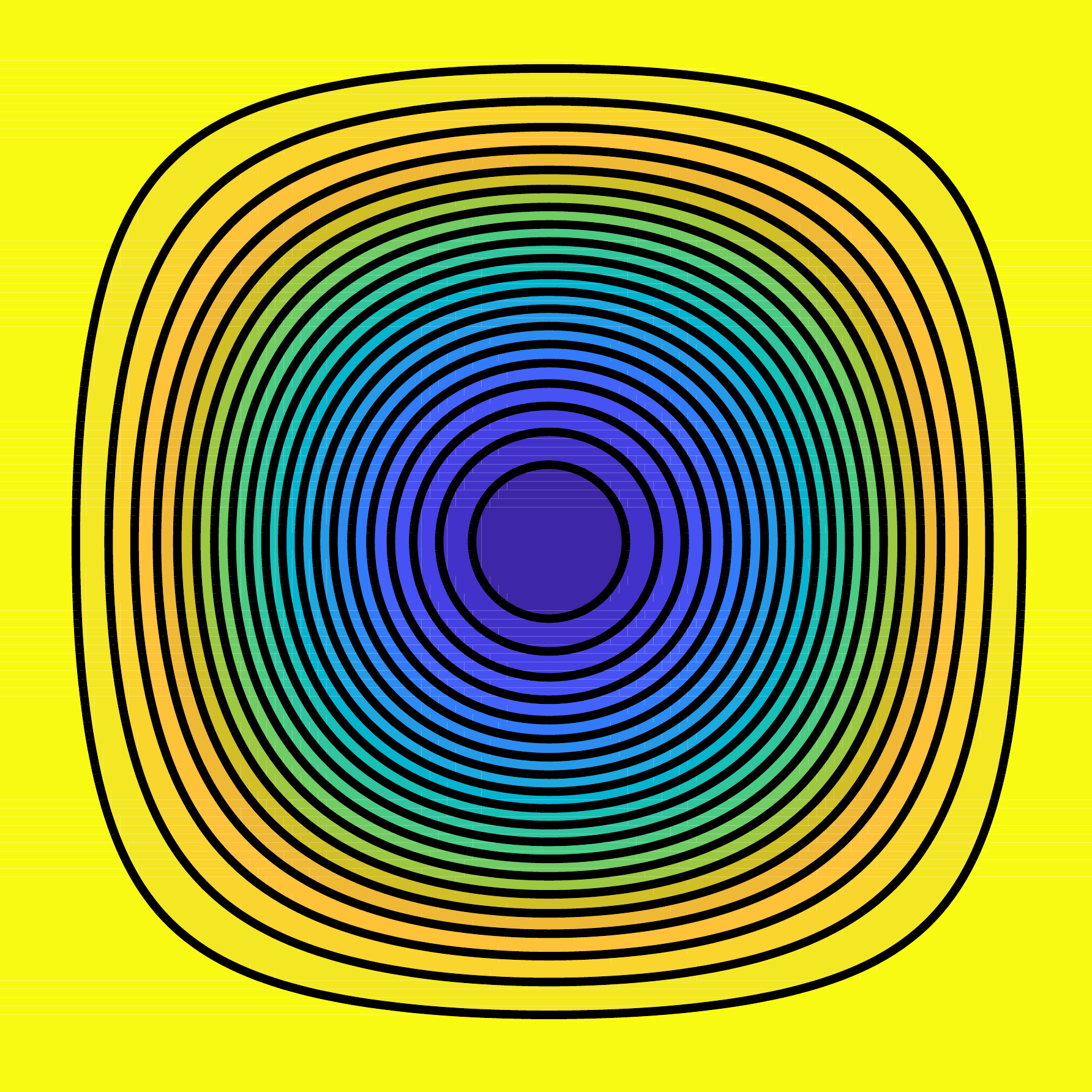}\label{fig:h4}}
	\caption{Isotropy of Laplace kernels shown with Fourier spectral magnitude.}
	\label{fig:kernels}
\end{figure}

Let us represent the Laplace operator with a convolution kernel, i.e.\ $\Delta U=k\ast U$. 
We compare the following four discrete convolution kernels as promising Laplace operator options for 2D images:
\begin{alignat}{2}
k_1 &= \left[ \begin{array}{ccc} \frac{1}{8} & \frac{1}{8} & \frac{1}{8} \\
\frac{1}{8} & -1 & \frac{1}{8} \\
\frac{1}{8} & \frac{1}{8} & \frac{1}{8} 
\end{array} \right]\!,\,&
k_2 &= \left[ \begin{array}{ccc} -\frac{1}{16} & \frac{5}{16} & -\frac{1}{16} \\
\frac{5}{16} & -1 & \frac{5}{16} \\
-\frac{1}{16} & \frac{5}{16} & -\frac{1}{16} 
\end{array} \right],\\
k_3 &= \left[ \begin{array}{ccc} \frac{1}{20} & \frac{1}{5} & \frac{1}{20} \\
\frac{1}{5} & -1 & \frac{1}{5} \\
\frac{1}{20} & \frac{1}{5} & \frac{1}{20} 
\end{array} \right]\!,\,&
k_4 &= \left[ \begin{array}{ccc} \frac{1}{12} & \frac{1}{6} & \frac{1}{12} \\
\frac{1}{6} & -1 & \frac{1}{6} \\
\frac{1}{12} & \frac{1}{6} & \frac{1}{12} 
\end{array} \right].
\label{eq:kernel3}
\end{alignat}
Kernel $k_1$ is a common isotropic Laplace operator. Kernel $k_2$ is from the mean curvature filter in~\cite{gong:cf}. Kernel $k_3$ originates from the numerical analysis field~\cite{patra2006stencils}. Kernel $k_4$ is common in the image processing community~\cite{isoLaplace:1999}.  To analyze the isotropy of these kernels, we transformed them into Fourier domain with their spectral magnitude plotted in Fig.~\ref{fig:kernels} with isolines. 

Kernel $k_4$ is seen to be the most isotropic one and is thus used in this work as the discrete Laplace operator.

\subsubsection{Discrete Normal Direction}
\begin{figure}[h]
	\centering
	\includegraphics[width=.7\linewidth]{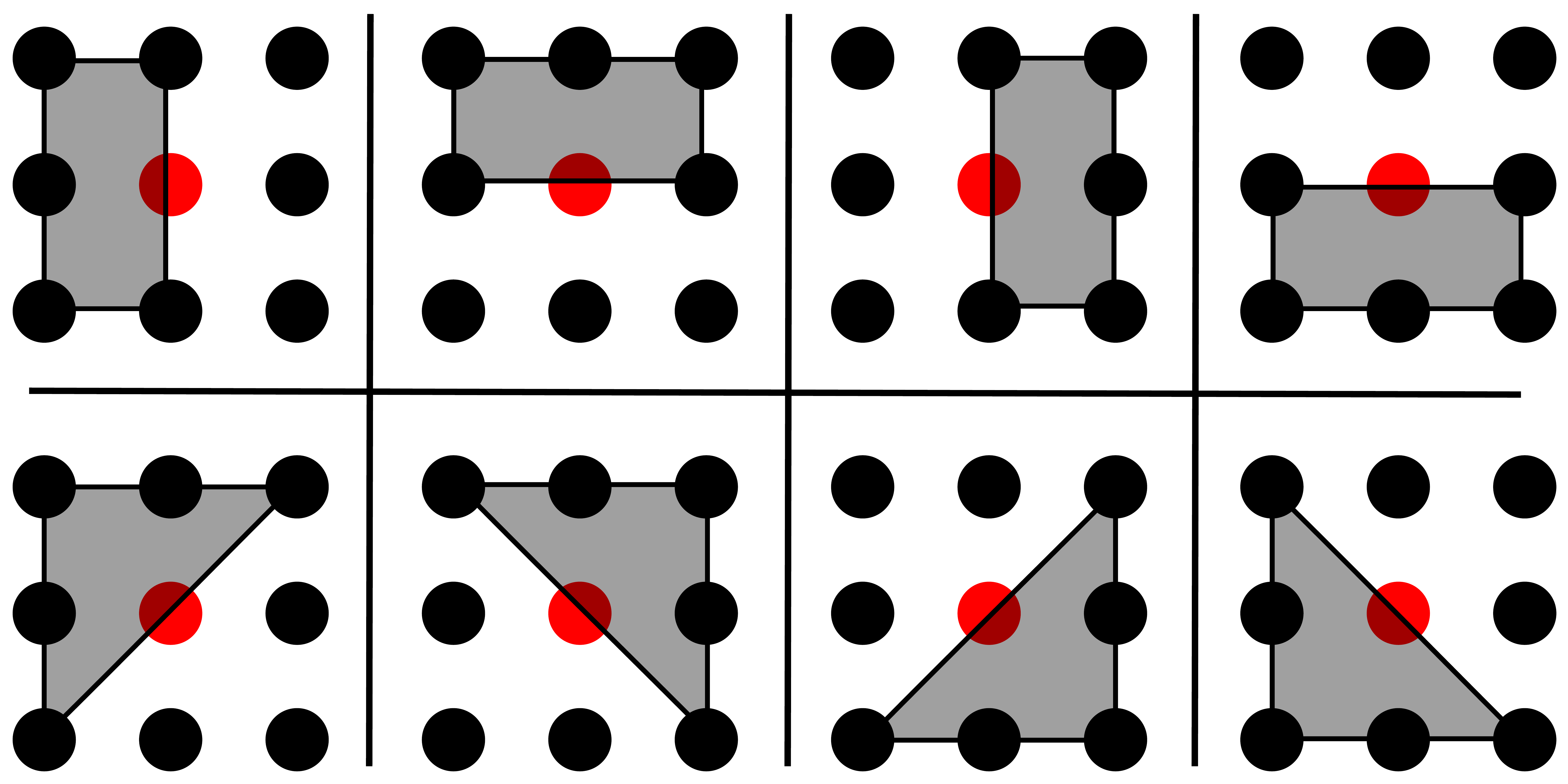}
	\caption{Eight cases $\{\vec{n}_i|i=1,..,8\}$ of half window regression}
	\label{fig:window}
\end{figure}

Since gradient, MC and WMC are local properties, as their support region gets smaller, they provide better approximation, especially at sharp edges (although this may differ in smoother regions).
In this paper, we choose a $3\times3$ window and consider only 8 possible normal directions separated by 45$^\circ$ indicated with half windows shown in Fig.~\ref{fig:window}. 
At any location $\vec{x}$, its normal $\vec{n}$ is then approximated by one of these eight cases $\{\vec{n}_i|i=1,..,8\}$. 
Note that these directions include the horizontal and vertical gradients typically used in (anisotropic) TV regularization.

The limitation of this discrete normal assumption is that the continuous normal $\vec{n}$ is only approximated by the discrete $\vec{n}_i$, with potential approximation errors.
Nevertheless, with a limited discrete directions as in Fig.~\ref{fig:window}, the normal diffusion term of $H^w$ in Eq.(\ref{eq:rhs}) becomes enumerable such that its computation is fast and straight-forward. 
Additionally, by replacing this normal diffusion term with regression of half windows, \emph{any numerical instability due to $\|\nabla U\|=0$ is avoided}. Moreover, since the diffusion can only happen along edges (but not normal to edges), this discretization also helps to preserve edges.  This can be envisioned as an extreme case of anisotropic diffusion.

For the half-window regressions, we use Neumann boundary conditions. 
This imposes the reflection symmetry property in our convolution kernels~\cite{gong2009symmetry}. 
This boundary condition is imposed at every pixel, which is different from traditional boundary conditions that are only enforced at large gradient locations. 

\subsection{Convolution Scheme}
\label{sec:kernels}
Combining the proposed discrete Laplace kernel with the 8 half-windows with Neumann boundary conditions~\cite{gong2009symmetry} yields the following eight convolution kernel candidates:
\begin{alignat*}{2}
h_1 &= \left[ \begin{array}{ccc} \frac{1}{6} & \frac{1}{6} & 0 \\
\frac{1}{3} & -1 & 0 \\
\frac{1}{6} & \frac{1}{6} & 0
\end{array} \right]\!,\,&
h_2 &= \left[ \begin{array}{ccc} \frac{1}{6} & \frac{1}{3} & \frac{1}{6} \\
\frac{1}{6} & -1 & \frac{1}{6} \\
0 & 0 &0 
\end{array} \right] \,,\\
h_3 &= \left[ \begin{array}{ccc} 0 & \frac{1}{6} & \frac{1}{6} \\
0 & -1 & \frac{1}{3} \\
0 & \frac{1}{6} & \frac{1}{6}
\end{array} \right]\!,\,&
h_4 &= \left[ \begin{array}{ccc} 0 & 0 & 0 \\
\frac{1}{6} & -1 & \frac{1}{6} \\
\frac{1}{6} & \frac{1}{3} &\frac{1}{6} 
\end{array} \right] \,,\\
h_5 &= \left[ \begin{array}{ccc} \frac{1}{6} & \frac{1}{3} & \frac{1}{12} \\
\frac{1}{3} & -1 & 0 \\
\frac{1}{12} & 0 & 0
\end{array} \right]\!,\,&
h_6 &= \left[ \begin{array}{ccc} \frac{1}{12} & \frac{1}{3} & \frac{1}{6} \\
0 & -1 & \frac{1}{3} \\
0 & 0 &\frac{1}{12} 
\end{array} \right],\\
h_7 &= \left[ \begin{array}{ccc}  0 & 0 & \frac{1}{12} \\
0 & -1 & \frac{1}{3} \\
\frac{1}{12} & \frac{1}{3} & \frac{1}{6}
\end{array} \right]\!,\,&
h_8 &= \left[ \begin{array}{ccc} \frac{1}{12}  & 0 & 0 \\
\frac{1}{3} & -1 & 0\\
\frac{1}{6} & \frac{1}{3} &\frac{1}{12} 
\end{array} \right]\!.
\end{alignat*}
From these kernels, we compute eight signed distances
\begin{equation}
\label{eq:conv}
	d_i=h_i\ast U\,,~\forall i=1,..,8\ .
\end{equation}

Resulting $\{d_i\}$ can be interpreted as the signed projection distances to the hyperplanes defined by each half-window. 
Therefore, the $d_i$ that has the minimum absolute value is the proximal projection distance that has the highest probability to a minimal surface. 
We use it as our estimation for WMC, i.e.
\begin{equation}
\label{eq:act}
	H^w\approx d_m,~~\mathrm{where}~~m=\arg\min_i\{|d_i|;\,i=1,..,8\}\,.
\end{equation} We call the algorithm described by Eq.~(\ref{eq:conv}) and (\ref{eq:act}) as Half Laplace Operator because the support region is a half window. 

\subsection{Neural Network Representation}
\begin{figure}[h]
    \centering
    {\includegraphics[width=.7\linewidth]{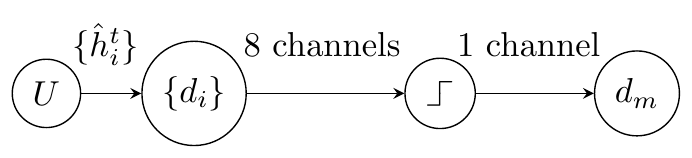}}
	\caption{Neural network representation of our scheme}
	\label{fig:scheme_net}
\end{figure}

This computation scheme has a convolutional neural network representation. 
The convolutions in Eq.~\ref{eq:conv} can be represented as convolution layers in a neural network structure while Eq.~\ref{eq:act} acts as a nonlinear activation function. 
Thus, our discrete computation scheme can be interpreted as a neural network.
For instance, an architecture for computing mean curvature is shown in Fig.~\ref{fig:scheme_net}. The step symbol in Fig.~\ref{fig:scheme_net} indicates Eq.~\ref{eq:act} and is the activation function in neural networks.   

Based on such neural-network representation, two further extensions of our discrete computation scheme can be envisioned.
First, $\{h_i\}$ could be extended as \emph{learnable} kernels $\{\hat{h}^t_i\}$ ($t$ is the layer index) for a specific dataset, such as natural images or medical images of a certain modality and given anatomy; such that these learned kernels would be more effective than the general $\{{h}_i\}$. 
Since $\{\hat{h}^t_i\}$ are coupled with the kernels $\{\hat{h}^{t-1}_i\}$ and $\{\hat{h}^{t+1}_i\}$ in the previous and next layers, their layer specific setting might be more efficient in minimizing the loss function. 
Second, the number of kernels can be increased, allowing for more than 8 \emph{normal} directions or possible operator structures. 
We present numerical results from a neural network representation of our computational scheme later in Section~\ref{sec:cnn}.

\subsection{Support Region and Computational Complexity}
The presented computation scheme has linear computational complexity and is computationally very efficient, with an implementation presented in the Appendix.
The eight convolution operations can also be implemented in parallel on modern hardwares, such as Graphic Processing Unit (GPU). 
Our parallel implementation can process 33.2\,Giga-pixels per second on a TITAN X Pascal GPU, using its native CUDA language. This indicates that our proposed methods can be used for images with very large sizes or in real-time image processing tasks. 
Furthermore, since the half windows in Fig.~\ref{fig:window} have overlapping regions, it is also possible to use integral images to further improve computational performance.

Since mean curvature is a local property, smaller support region is always preferred (as well resulting in a faster computation). 
Although the continuous definition of $H$ can represent edges from a mathematical point of view, its finite difference approximation in Eq.~\ref{eq:2dlong} requires a relatively larger support region ($5\times5$ or larger depending on the finite difference scheme used) and thus cannot preserve edges due to such larger support region. 
The support region for our proposed scheme has only 6 taps (half-windows seen in Fig.~\ref{fig:window}) and thus WMC $H^w$ can preserve edges better than MC $H$.
This is demonstrated in the next section.

\section{Experiments}

\subsection{Comparison between $\|\nabla U\|$, $H$ and $H^w$}
\begin{figure}
	\subfigure[original]{\includegraphics[width=.24\linewidth]{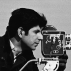}\label{fig:cam_patch}}
	\subfigure[$\|\nabla U\|_2$+90]{\includegraphics[width=.24\linewidth]{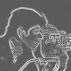}\label{fig:cam_patch_grad}}
	\subfigure[$H\times 20+128$]{\includegraphics[width=.24\linewidth]{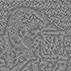}\label{fig:cam_patch_H}}
	\subfigure[$H^w\times 2+128$]{\includegraphics[width=.24\linewidth]{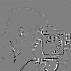}\label{fig:cam_patch_HW}}
	\caption{A close-up of (a) an image; (b) its gradient norm, (c) mean curvature, and (d) WMC; all scaled to [0,255] gray value range as shown.}
	\label{fig:perform}
\end{figure}

In order to show the difference between gradient $\|\nabla U\|$, mean-curvature $H$, and WMC $H^w$, we apply them on a sample image, where a patch is shown in Fig.~\ref{fig:perform} as a close-up. 
Since WMC has a smaller support region, it captures local geometries better than MC; e.g.\ the variance of $H^w$ is small in the sky, while $H$ still presents strong local variance in Fig.~\ref{fig:perform}.

\subsection{Comparison of ${\cal R}_{TV}$, ${\cal R}_{H}$, and ${\cal R}_{\mathrm{area}}$ }
\label{sec:compare}

\begin{figure*}
	\centering
	\begin{subfigure}[original]
		{\includegraphics[width=0.22\linewidth,height=0.22\linewidth]{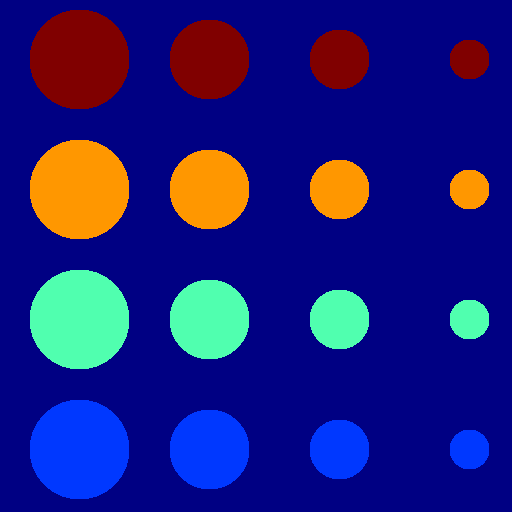}}
		{\includegraphics[height=0.22\linewidth]{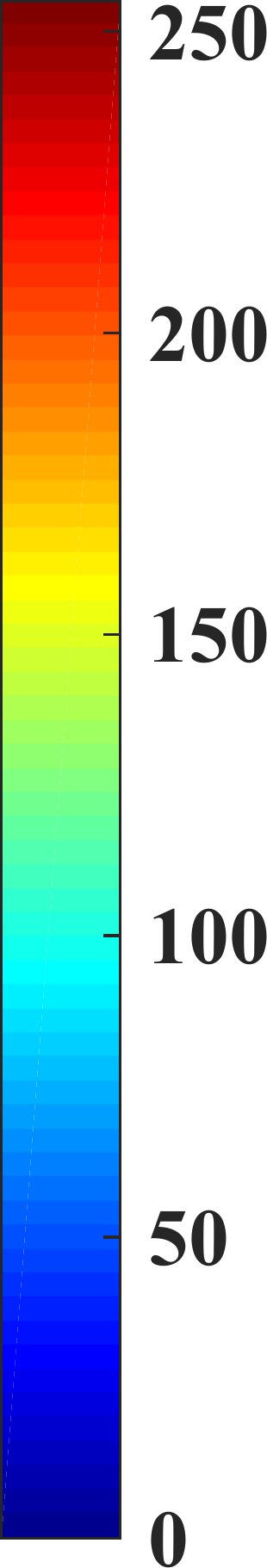}}
		\label{fig:pwmrf} 
	\end{subfigure} \vline width 2pt~~~
\begin{subfigure}[${\cal R}_{TV}$ with $\lambda=300$]
	{\includegraphics[width=0.22\linewidth,height=0.22\linewidth]{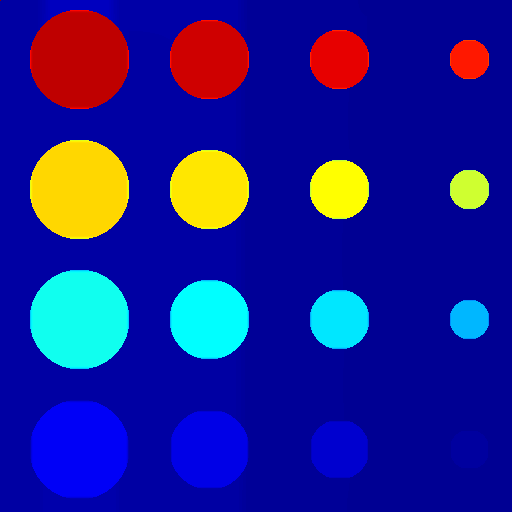}}
\end{subfigure}
\begin{subfigure}[${\cal R}_{TV}$ with $\lambda=700$]
	{\includegraphics[width=0.22\linewidth,height=0.22\linewidth]{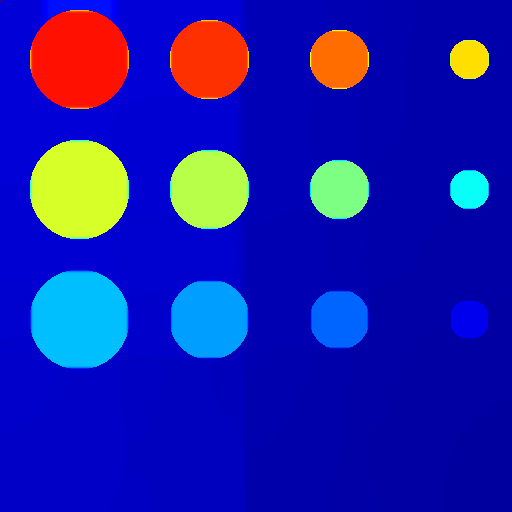}}
\end{subfigure}
\begin{subfigure}[${\cal R}_{TV}$ with $\lambda=1000$]
	{\includegraphics[width=0.22\linewidth,height=0.22\linewidth]{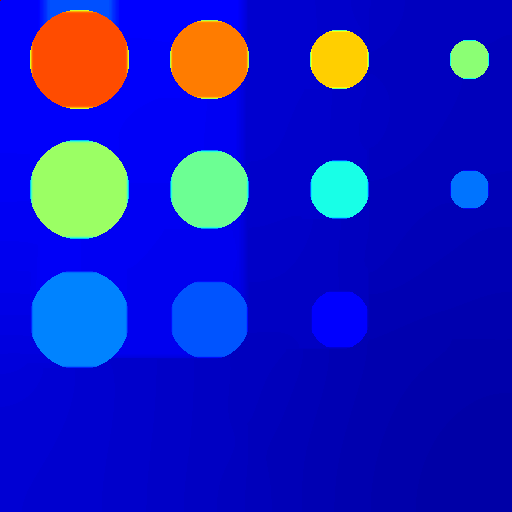}}
\end{subfigure}
	\begin{subfigure}[one column profile (contrast behavior)]
		{\includegraphics[width=0.27\linewidth]{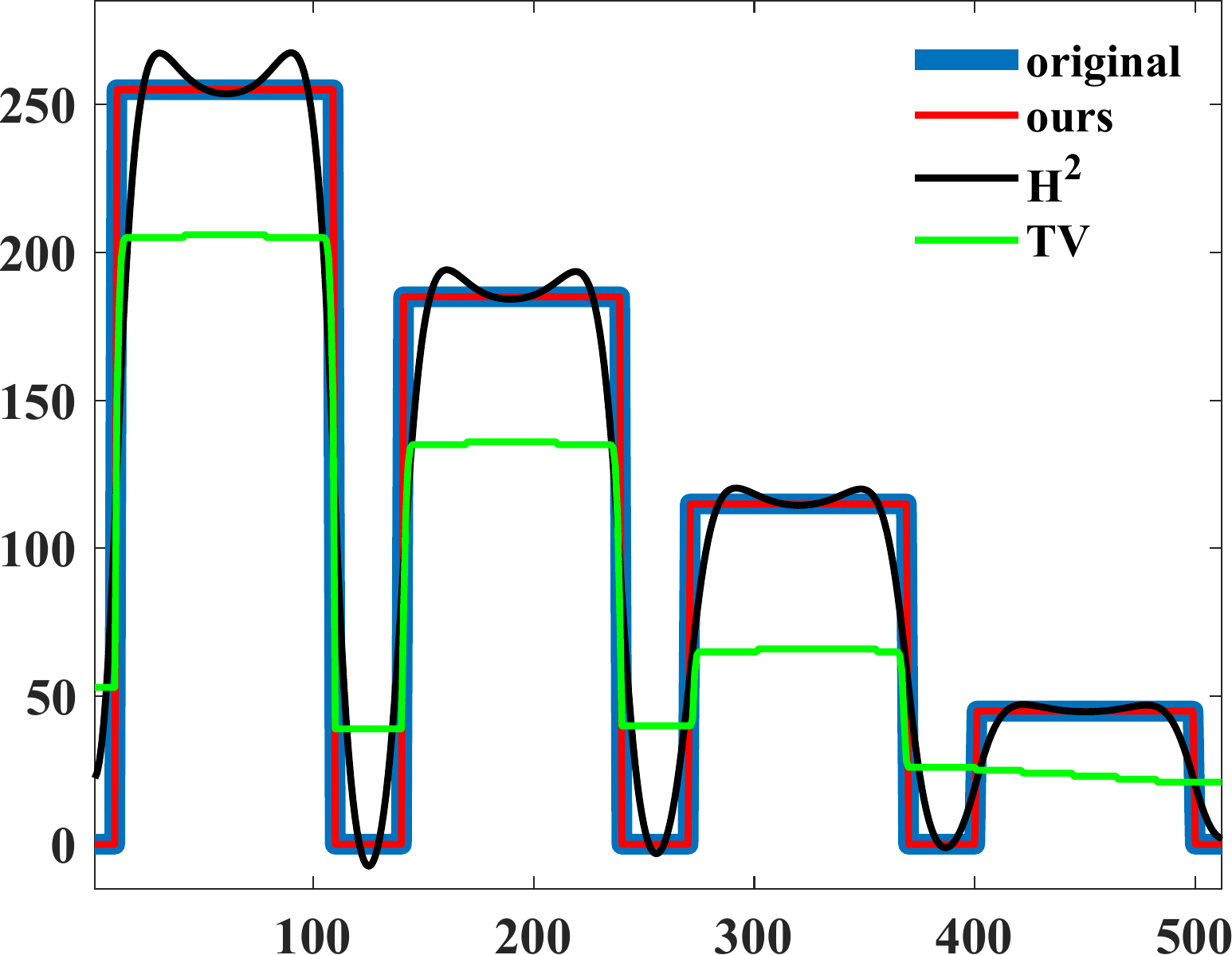}}
	\end{subfigure}\vline width 2pt~~~\,
	\begin{subfigure}[${\cal R}_{H}$ with $\lambda=20$]
		{\includegraphics[width=0.22\linewidth,height=0.22\linewidth]{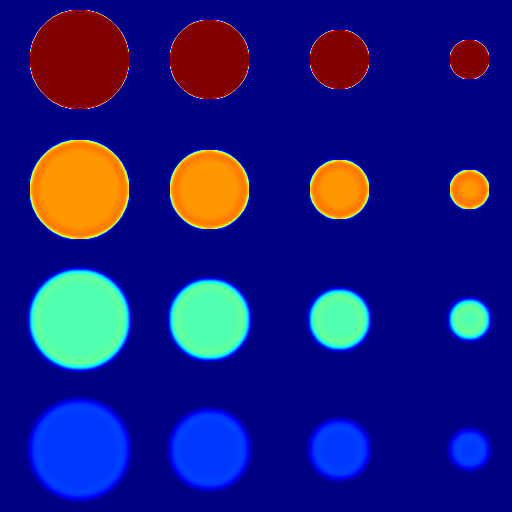}}
	\end{subfigure}
	\begin{subfigure}[${\cal R}_{H}$ with $\lambda=150$]
		{\includegraphics[width=0.22\linewidth,height=0.22\linewidth]{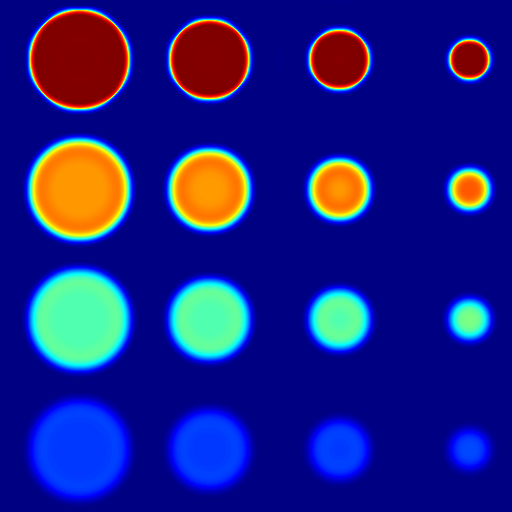}}
	\end{subfigure}
	\begin{subfigure}[${\cal R}_{H}$ with $\lambda=1000$]
		{\includegraphics[width=0.22\linewidth,height=0.22\linewidth]{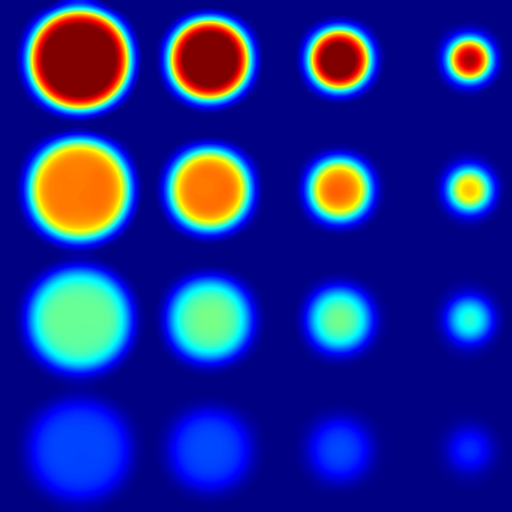}}
	\end{subfigure}
	\begin{subfigure}[one row profile (scale behavior)]
		{\includegraphics[width=0.27\linewidth]{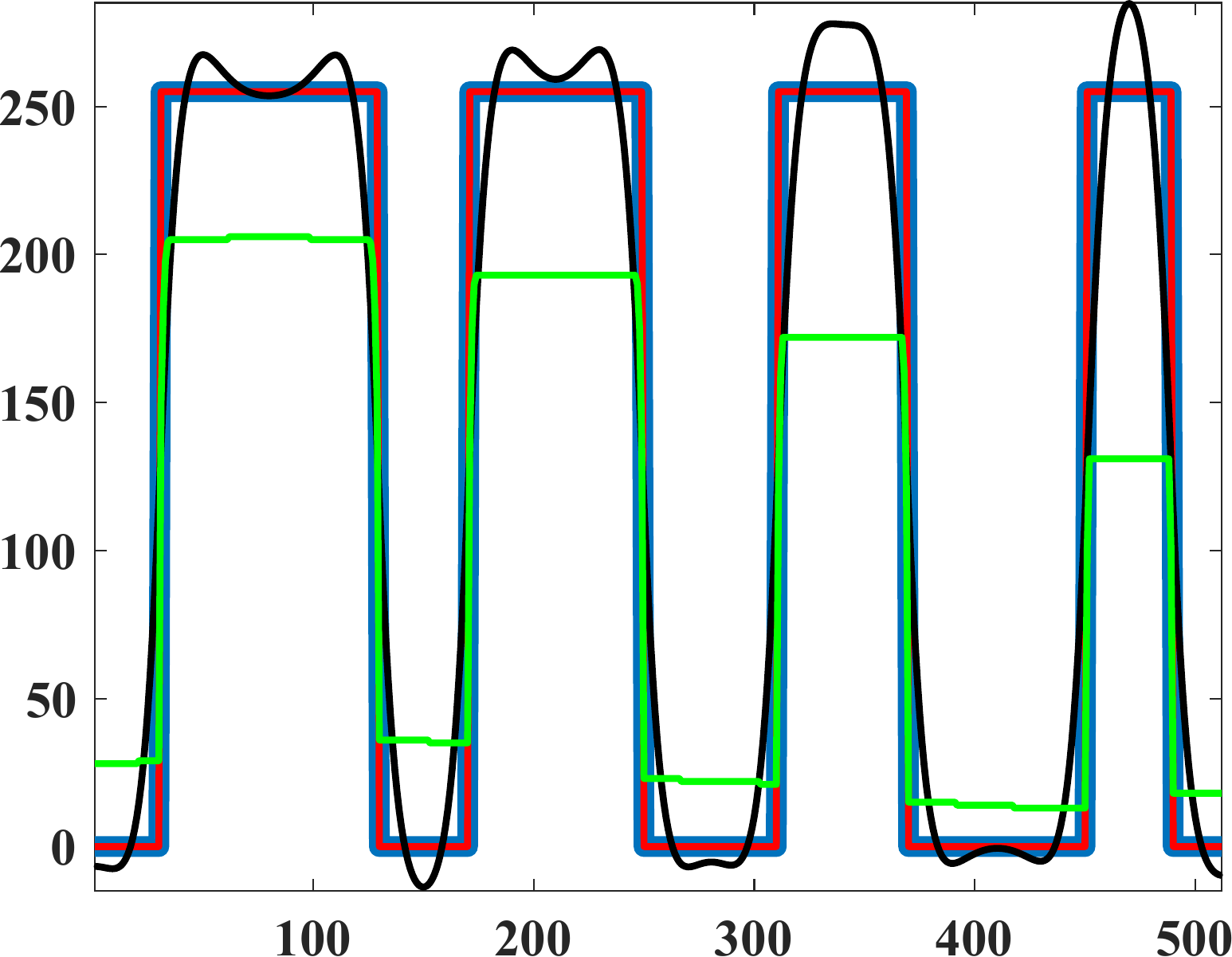}}
	\end{subfigure}\vline width 2pt
	~~~
	\begin{subfigure}[${\cal R}_{\mathrm{Area}}$ with $\lambda=20$]
		{\includegraphics[width=0.22\linewidth,height=0.22\linewidth]{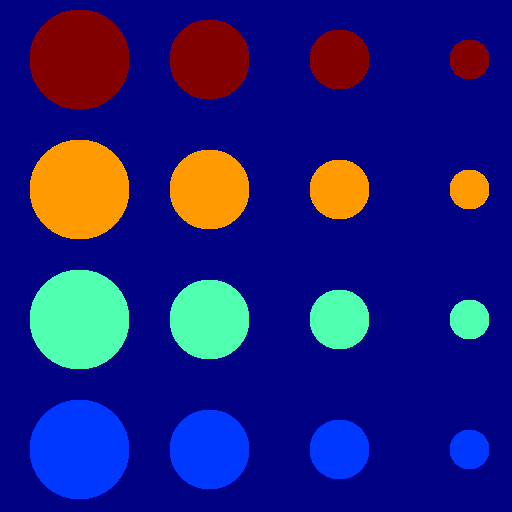}}
	\end{subfigure}
	\begin{subfigure}[${\cal R}_{\mathrm{Area}}$ with $\lambda=150$]
		{\includegraphics[width=0.22\linewidth,height=0.22\linewidth]{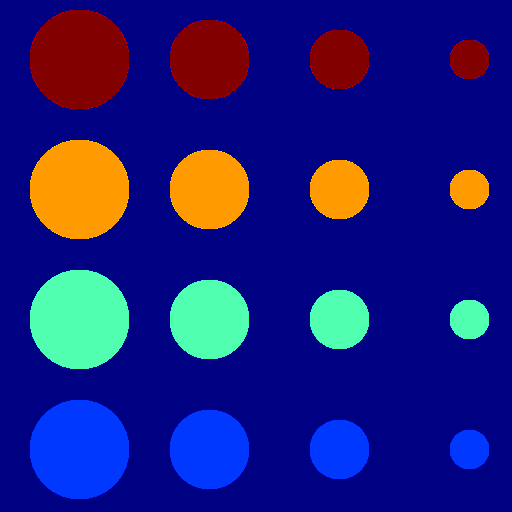}}
	\end{subfigure}
	\begin{subfigure}[${\cal R}_{\mathrm{Area}}$ with $\lambda=200$]
		{\includegraphics[width=0.22\linewidth,height=0.22\linewidth]{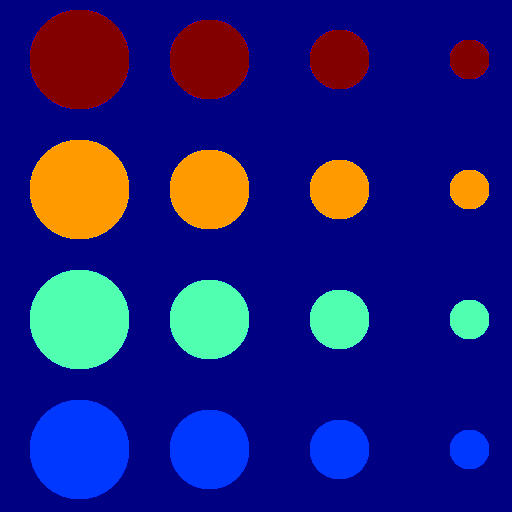}}
	\end{subfigure}
	\caption{Comparison of ${\cal R}_{TV}$, ${\cal R}_{H}$ and ${\cal R}_{\mathrm{Area}}$. 
	(a) original image; (b), (c), and (d) are results from TV regularization; (f), (g), and (h) are results from mean curvature regularization; (j), (k), and (l) are results from area regularization. (e) and (l) show the cross-sectional profiles for the first column and first row inclusions, respectively, in (d), (h), and (l).}
	\label{fig:3R}
\end{figure*}

We compare three different regularization options for the same imaging problem
\begin{equation}
U^\star=\arg\min_U\left\{\frac{1}{2}\|AU-f\|_2^2+\lambda{\cal R}(U)\right\}\,,
\end{equation}
where ${\cal R}$ is one of ${\cal R}_{TV}$, ${\cal R}_{H}$ and ${\cal R}_{\mathrm{area}}$. 
For simplicity, we set $A$ as identity matrix and checked the behaviors of these regularizations at different $\lambda$. 
We used a numerical ground-truth image seen in Fig.~\ref{fig:3R}(a), where each row of circles indicate scale change and each column indicate contrast difference.  

For ${\cal R}_{TV}$ case, we solve it using~\cite{jia2010fast} with an arbitrarily high 50\,K iterations. We set $\lambda=300$, 700, and 1000 with the resulting $U^\star$ seen in Fig.~\ref{fig:3R}(b-d).  
Herein relatively large values of $\lambda$ were selected to demonstrate the effect of and artifacts from regularization at its extreme. 
When $\lambda$ is increased, staircase artifacts are observed and the contrast is lost, where the smaller circles with lower contrast get smoothed out first. 
Note that although the TV model can capture sharp edges ideally~\cite{OsherTV2005}, it may lose contrast, especially at high regularization weights, as seen in cross-sections in Fig.~\ref{fig:3R}(e and i).

For ${\cal R}_{H}$ case, we solve it using~\cite{Carlos:2010}. We set $\lambda=20$, 150, and 1000, with the results seen in Fig.~\ref{fig:3R}(f-h). 
When $\lambda$ is increased, resulting images get smoother. 
Since mean-curvature $H$ is contrast invariant, the results are robust to contrast change, but not robust to scale change.

For area regularization, we use gradient descent for the data fitting term with our approximation Eq.~\ref{eq:area_reg} for ${\cal R}_{\mathrm{area}}$. 
More specifically, we apply the following iteration
\begin{equation}
\frac{\partial U}{\partial t}=-A^T(AU-f)-\lambda\frac{\partial {\cal R}}{\partial U}\,,
\end{equation} 
where $-\frac{\partial {\cal R}_{\mathrm{area}}}{\partial U}\approx H^w(U)$. We set $\lambda=20$, 150, and 200, with the results seen in Fig.~\ref{fig:3R}(j-l). 
Note that with increasing $\lambda$, the shrinking and smoothing behavior is the same across all scale and contrast levels. 
This empirical observation corroborates the contrast invariance discussion of area regularization earlier in Section~\ref{sec:grad}.
Since the normal directions are limited to 8 possible cases, WMC results at high regularization weights might get axis-biased; nonetheless this artifact is not distracting nor often observable in our results, as demonstrated in the next sections.

Fig.~\ref{fig:3R}(e) shows the contrast profiles for the three methods above along the first column of circles, and similarly Fig.~\ref{fig:3R}(i) along the first row of circles. 
Our WMC results are seen to be the closest to the ground-truth.   

\subsection{Data Fidelity with $\ell_1$ Norm}
\begin{figure*}[!htb]
	\centering
	\subfigure[original input $f$ ]{\includegraphics[width=.19\linewidth]{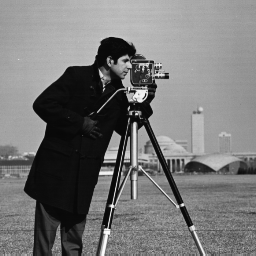}}
	\subfigure[$\lambda$=1, SSIM=0.9106  ]{\includegraphics[width=.19\linewidth]{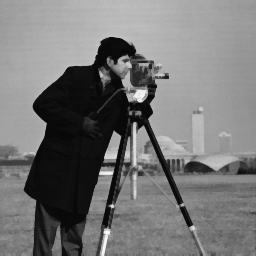}}
	\subfigure[$\lambda$=3, SSIM=0.8610 ]{\includegraphics[width=.19\linewidth]{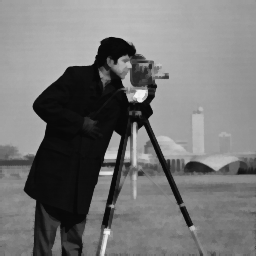}}
	\subfigure[$\lambda$=10, SSIM=0.8171 ]{\includegraphics[width=.19\linewidth]{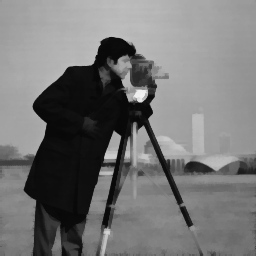}}
	\subfigure[$\lambda$=20, SSIM=0.8000 ]{\includegraphics[width=.19\linewidth]{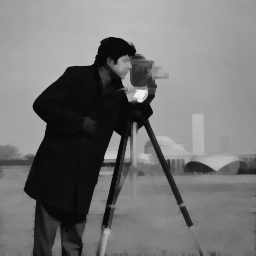}}
	
	\hspace{.19\linewidth}
	\subfigure[$f-U$+128 when $\lambda$=1  ]{\includegraphics[width=.19\linewidth]{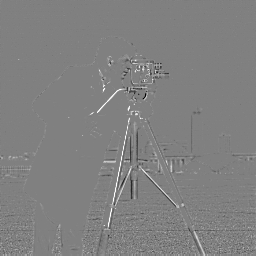}}
	\subfigure[$f-U$+128 when $\lambda$=3 ]{\includegraphics[width=.19\linewidth]{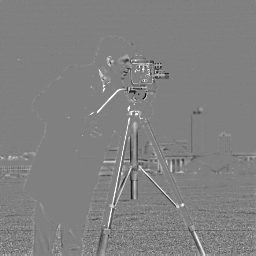}}
	\subfigure[$f-U$+128 when $\lambda$=10 ]{\includegraphics[width=.19\linewidth]{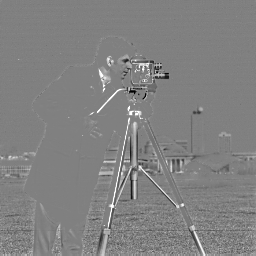}}
	\subfigure[$f-U$+128 when $\lambda$=20 ]{\includegraphics[width=.19\linewidth]{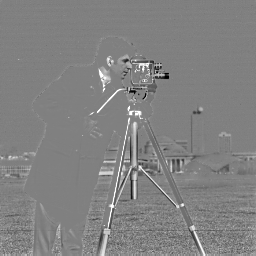}}
	\caption{The effect of increasing $\lambda$, where the second row shows the residual $f-U$.  Qualitatively speaking, with $\lambda=1$ the handle of the camera is still visible, while with $\lambda=3$ the handle disappears.
		With $\lambda=10$ the roof of the tall building is no more visible. 
		With $\lambda=20$ the results do not further change significantly, while major structures are still preserved to a large extent.}
	\label{fig:exp_L1_lambda}
\end{figure*}

We herein study the following $\ell_1$-based variation model, which is known to treat Laplace noise well and is robust to outliers while imposing sparsity:
\begin{equation}
	\min_U\left\{\|AU-f\|_1+\lambda {\cal R}(U)\right\}\,.
\end{equation}
We show the solution of this model based on our WMC-based area regularization, using the following primal-dual method.

For $b=AU-f$, we then have the following model
\begin{equation}
\min_U\left\{\|b\|_1+\lambda {\cal R}(U)\right\}~s.t.~b=AU-f\,.
\end{equation} 
This is equivalent to the dual representation
\begin{equation}
\min_U\left\{\|b\|_1+\lambda {\cal R}(U)+\frac{1}{2\alpha}\|b-AU+f-d\|_2^2\right\}\,,
\end{equation} 
where $\alpha>0$ is a small constant and $d$ is a scaled dual variable.

Within an iterative solution, at each iteration, given $r \equiv AU^t-f+d^t$ there is a closed-form solution for $b$ as
\begin{eqnarray}
b=
\begin{cases}
r-\alpha, &r>\alpha\cr 0, &|r|\le \alpha \cr r+\alpha, &r<-\alpha\end{cases}\,,
\end{eqnarray}
and for $d$ as
\begin{eqnarray}
d^{t+1}=d^t+AU^t-f-b=
\begin{cases}
\alpha, &r>\alpha\cr r, &|r|\le \alpha \cr -\alpha, &r<-\alpha\end{cases}\,.
\end{eqnarray}
For $U$, the following problem needs to be solved
\begin{equation}
\min_U\left\{\lambda {\cal R}(U)+\frac{1}{2\alpha}\|b-AU+f-d^{t+1}\|_2^2\right\}\,.
\end{equation} 
for which one way is to use gradient descent as follows:
\begin{equation}
\begin{split}
\frac{\partial U}{\partial t}&=-\lambda\frac{\partial {\cal R}}{\partial U}-\frac{1}{\alpha}A^T(AU^t-f+d^{t+1}-b)\\
&=-\lambda\frac{\partial {\cal R}}{\partial U}-\frac{1}{\alpha}A^T(2d^{t+1}-d^t)\,.
\end{split}
\end{equation} 
As a result, $b$ is eliminated from the optimization procedure, $d$ and $U$ need to be updated alternately until convergence. 
As only a matrix multiplication is needed each for $d^{t+1}$ and $U^{t+1}$ updates, the iterations can be computed very fast.

For area regularization as in Eq.~\ref{eq:area_reg}, we then have
\begin{equation}
\frac{\partial U}{\partial t}\approx\lambda H^w(U^t)-\frac{1}{\alpha}A^T(2d^{t+1}-d^t)\,.
\end{equation}
We can solve this equation using our computation scheme for $H^w$.
When $A$ is the identity matrix, this model leads to a smoothing problem. 
Results of this model for different $\lambda$ are shown in Fig.~\ref{fig:exp_L1_lambda} with a qualitative assessment on the preservation of most major structures even at higher $\lambda$. We used structural similarity index measurement (SSIM) to quantify the structural similarity between images.

\subsection{Mean Curvature Flow with Edge Preservation}
\label{sec:flow}
\begin{figure*}
	\centering
	\subfigure[original ]{\includegraphics[width=.24\linewidth]{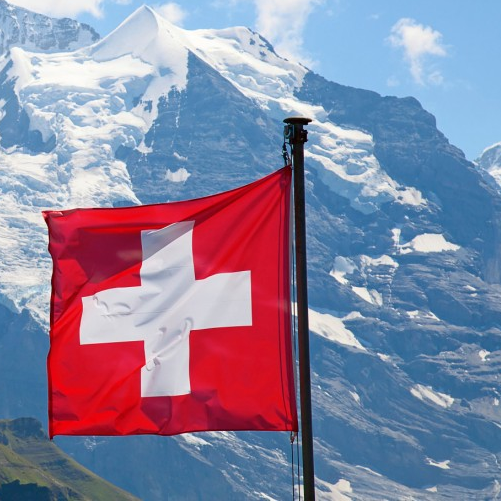}}
	\subfigure[deriv, t=10: SSIM=0.923  ]{\includegraphics[width=.24\linewidth]{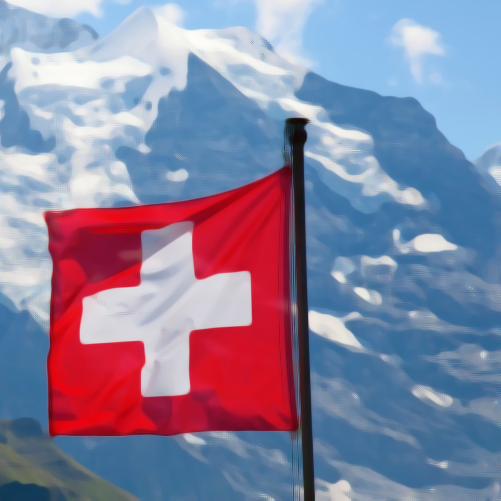}}
	\subfigure[deriv, t=50: SSIM=0.870  ]{\includegraphics[width=.24\linewidth]{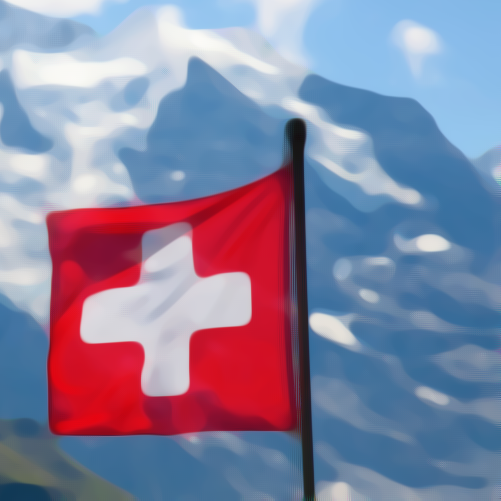}}
	\subfigure[deriv, t=200: SSIM=0.821 ]{\includegraphics[width=.24\linewidth]{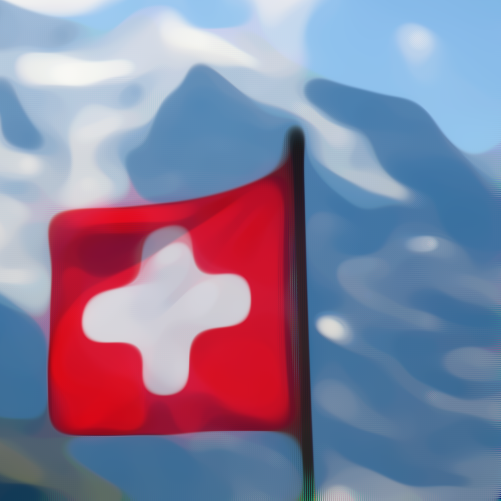}}
	\subfigure[patch details]{
		 \begin{overpic}[width=0.235\linewidth]{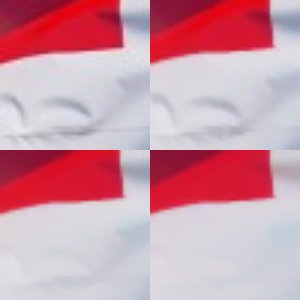}
			\put(20,85){\bf a}
			\put(70,85){\bf f}
			\put(20,36){\bf g}
			\put(70,36){\bf h}
		\end{overpic}
	}
	\subfigure[HalfLaplace, t=10: SSIM=0.974  ]{\includegraphics[width=.237\linewidth]{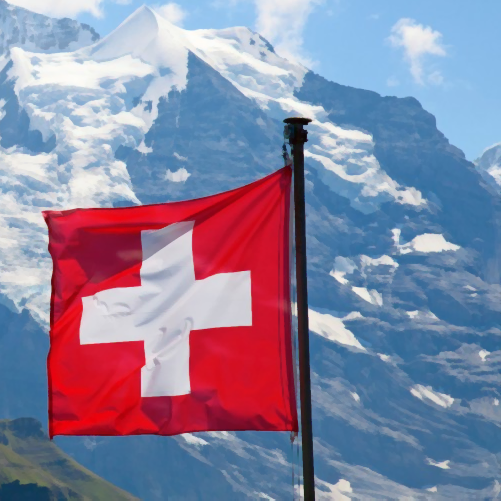}}
	\subfigure[HalfLaplace, t=$10^2$: SSIM=0.940   ]{\includegraphics[width=.237\linewidth]{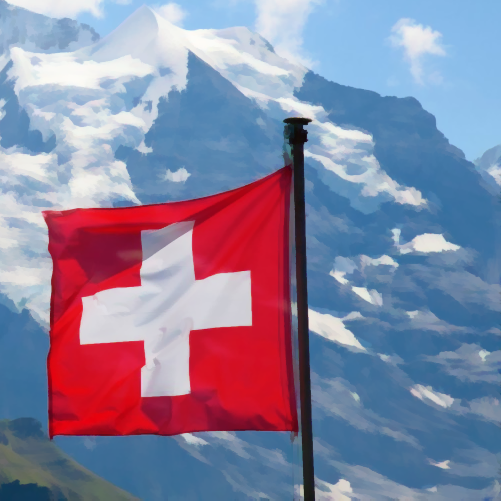}}
	\subfigure[HalfLaplace, t=$10^3$: SSIM=0.909   ]{\includegraphics[width=.237\linewidth]{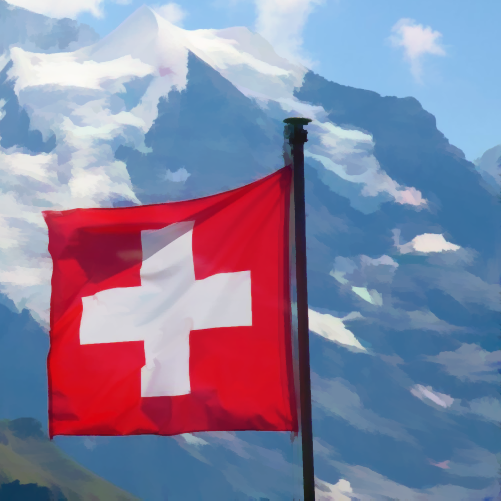}}
	\caption{Comparison of Mean Curvature Flow with the standard computation scheme (first row) and the proposed computation scheme (second row). Detailed patches from our method are shown in (e), where original is at the left up, t=10 at the right up, t=100 at left bottom and t=1000 at the right bottom. Our scheme preserves the edges and sharp corners (cf. the flag and the cross). Mean curvature flow was performed on each color channel separately.}
	\label{fig:exp_flow}
\end{figure*}

Given the advantages of WMC, we herein show that its proposed computation scheme is ideal, yielding significantly further advantages in its applications.
We study the iterative mean curvature flow application in Eq.(\ref{eq:MCF}), which is independent of an imaging modal and data term.
We discretize WMC using the conventional method or our proposed computation scheme, and present results at different iterations $t$ in Fig.~\ref{fig:exp_flow}.
As seen in Fig.~\ref{fig:exp_flow}, our discrete computation scheme can preserve local geometry and sharp edges much better compared to the traditional discretization. 
This is due to two reasons: (i) our method is performed within a smaller support with limited diffusion directions; and (ii) our computation does not require the image to be second-order differentiable, whereas the traditional discretization computes second derivatives hence requiring the result to be at least second-order smooth within given supports.

A close-up is shown in Fig.~\ref{fig:exp_flow}(e) to demonstrate the smoothing-while-edge-preserving behavior of our computation scheme. 
With increasing iteration number, the wrinkles on the flag are correctly smoothed out, while the edges of cross are well preserved.
A similar optimal behavior is also observed on the mountain and snow textures. 
In contrast, with traditional mean-curvature computation, the sharp corners become rounded while most structural information is also smoothed as Fig.~\ref{fig:exp_flow}(d). 

Note that the above is merely a difference from the two approximations (computation schemes) of the same WMC operation. 
We conclude that our proposed discrete computation is potentially more suitable for most image processing tasks.  

\subsection{Convolutional Neural Network Implementation}
\label{sec:cnn}

\begin{figure}
    \centering
{\includegraphics[width=.6\linewidth]{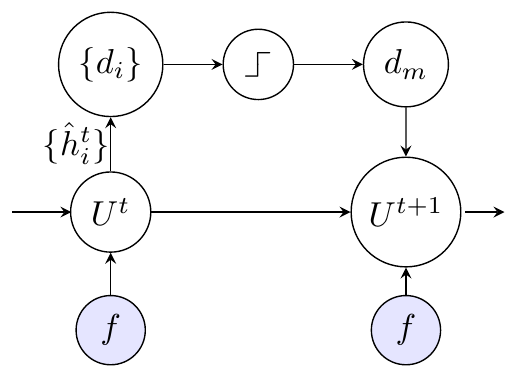}}
	\caption{Neural network implementation of variational model in Eq.~(\ref{eq:exp_model})}
	\label{fig:net}
\end{figure}

As mentioned, our computation scheme has a neural network representation, which can be used to find data adaptive kernels $\{\hat{h}_i\}$. 
Using our neural network structure in Fig.~\ref{fig:net}, we solve the following model
\begin{equation}
\label{eq:exp_model}
\min_U\left\{\frac{1}{s}\|AU-f\|_s^s+\lambda{\cal R}_{\mathrm{area}}(U)\right\}\,,
\end{equation} 
for $s=1$ and $s=2$. 
For simplicity, we again assume $A$ as identity. 
Traditional gradient decent method for this model required hundreds of iterations to converge, whereas our neural network method was able to obtain acceptable results with only two layers and eight learned filters. 
Neural network implementation thus required two to three orders of magnitude shorter time than the iteration method, when both are performed on the same hardware.

\begin{figure}
	\centering
	\includegraphics[width=\linewidth]{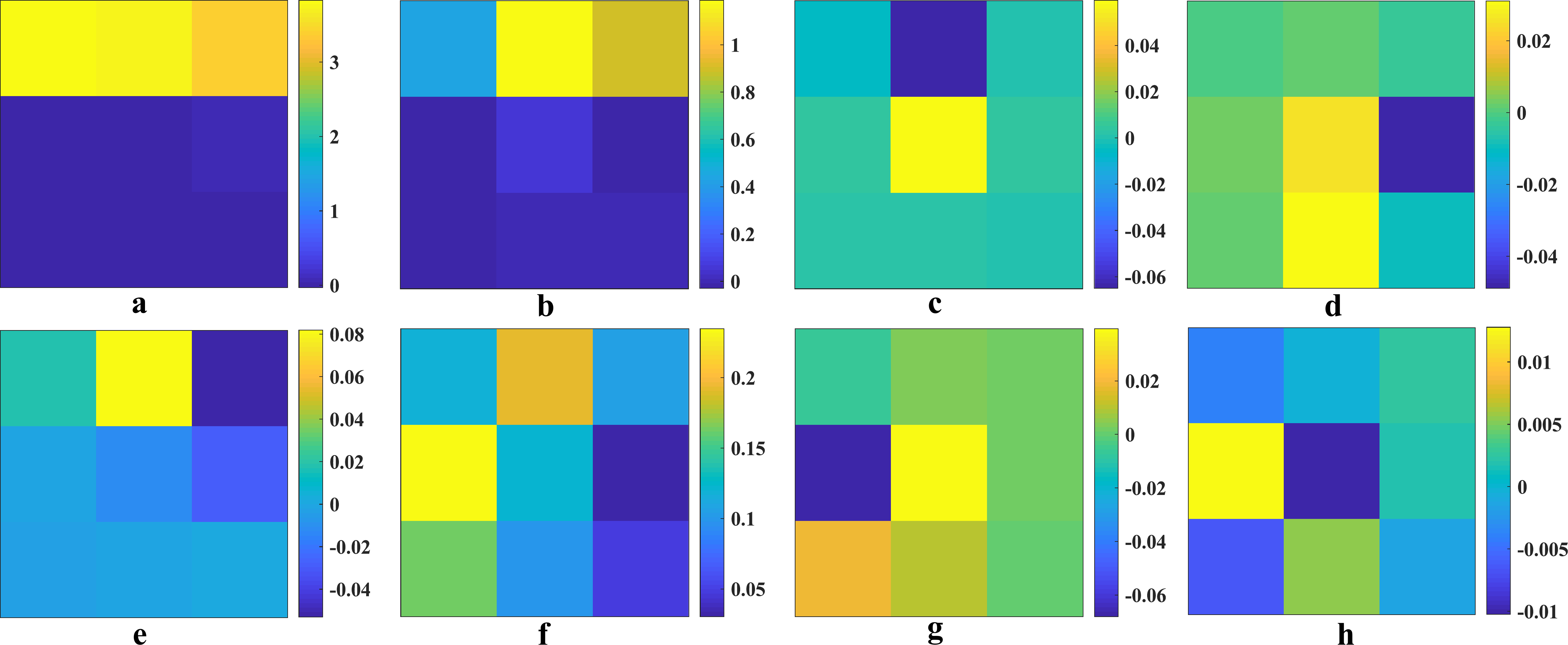}
	\caption{Learned kernels $\{\hat{h}^t_i\}$ in the first layer. Some of these kernels already look like half windows (e.g., a, b, e, g, h) and gradient operators (c, d).}
	\label{fig:learned}
\end{figure}

To obtain the learned kernels, we extracted 10,000 image patches of $128\times128$ pixels each, randomly from natural image dataset BSDS500. 
We used the network structure shown in Fig.~\ref{fig:net}, with two such consecutive layers (more layers did not reduce the loss function anymore for this problem). 
We separately learned 4, 8, and 16 convolutional filters of $3\times 3$ size each. We set a learning rate of $2\times10^{-4}$, batch size of 32, and regularization $\lambda$ of 5. The eight learned kernels in the first layer from our 8 filter network are shown in Fig.~\ref{fig:learned}. Some of these kernels already look like our half Laplace kernels.
The average loss function for different number of learned filters is tabulated in Table~\ref{Table:energy}, in comparison to the iterative method with 1000 iterations.
Note that merely a two-layer network yields better results (energy-level and quality) compared to the iterative approach, since these filters are now data-adaptive.
Since the filters $\{\hat{h}^t_i\}$ are independent from image resolution, they can be trained on low resolution images. 
Note that the content of testing images should be similar with the training dataset, i.e. natural scenery in this case.
Nevertheless, while these learned filters would be valid only for training-like input, the generic kernels given in Section \ref{sec:kernels} are valid for any input data. 
\begin{table}
	\centering
	\caption{Average energy level for different solvers.}
	\label{Table:energy}
	\begin{tabular}{c |c c c || c} %
		\hline
		\multicolumn{4}{c||}{Our Neural Network Method} &
		Iterative Method \\
		\hline
		& \# filters\,=\,4& 8 & 16 &  iteration=1000\\
		\hline
		$s=1$ & 35.0 & 33.3  & 32.7 & 39.4 \\
		\hline
		$s=2$ & 66.5  & 66.3 &  66.1 & 69.8\\  
		\hline
	\end{tabular}
\end{table}

\begin{figure*}
	\centering
	\subfigure[underwater]{\includegraphics[width=.32\linewidth]{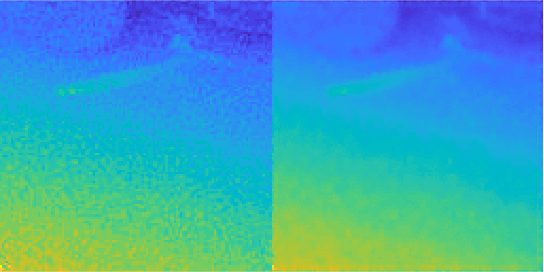}}
	\subfigure[elephant]{\includegraphics[width=.32\linewidth]{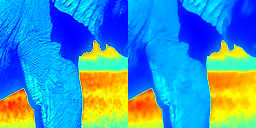}}
	\subfigure[tree branches]{\includegraphics[width=.32\linewidth]{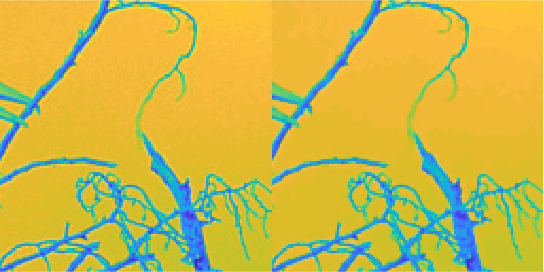}}
	\caption{In each sub-image, a close-up of (left) the original image and (right) the output from our neural network implementation with $s$$=$$2$, eight kernel filters, $\lambda$$=$$5$ are shown. 
		(a)~Image is smoothed while preserving edges. (b)~Wrinkles on the skin are removed, while elephant outline is well preserved. (c)~Contrast is well preserved while smoothing the background and the branches.}
	\label{fig:exp_L2_nn}
\end{figure*}

Figure~\ref{fig:exp_L2_nn} shows some example results from our neural network implementation with $s=2$, $\lambda=5$, and 8 kernels used. 
The details are seen to be removed, while the main structures being successfully preserved.  
With TensorFlow library and Python implementation, each prediction took 1.2\,ms on a GeForce 940MX (384 CUDA cores) with Windows 10 (Thinkpad T480) and 0.25\,ms on a GeForce GTX TITAN X (3072 CUDA cores) with Linux. Better performance can be obtained by using native CUDA C++ language.
A two-layer network is seen to be orders of magnitude faster than the iterative structure.
\begin{figure*}
	\centering
	\includegraphics[width=0.79\textwidth]{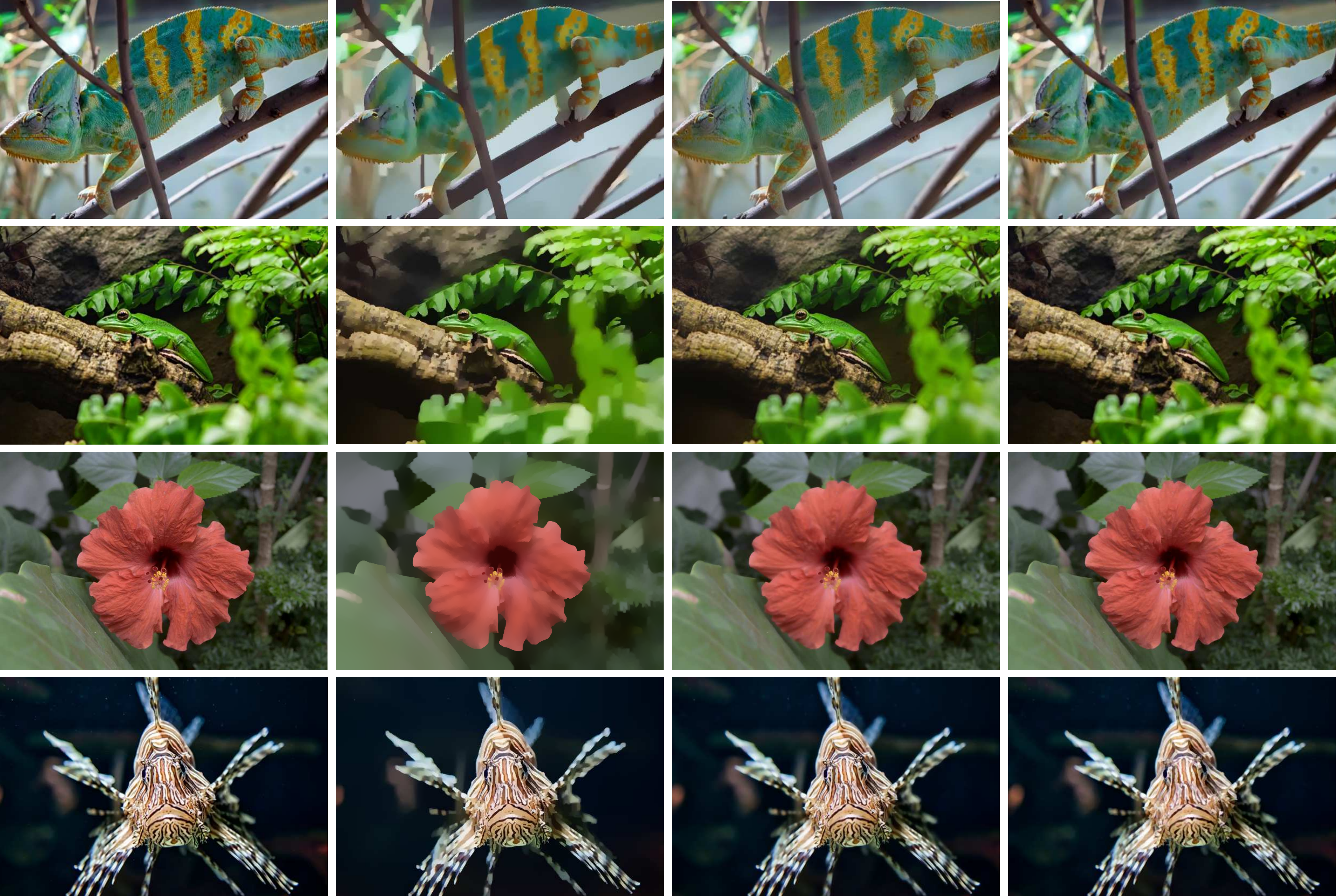}
	\includegraphics[width=0.194\textwidth]{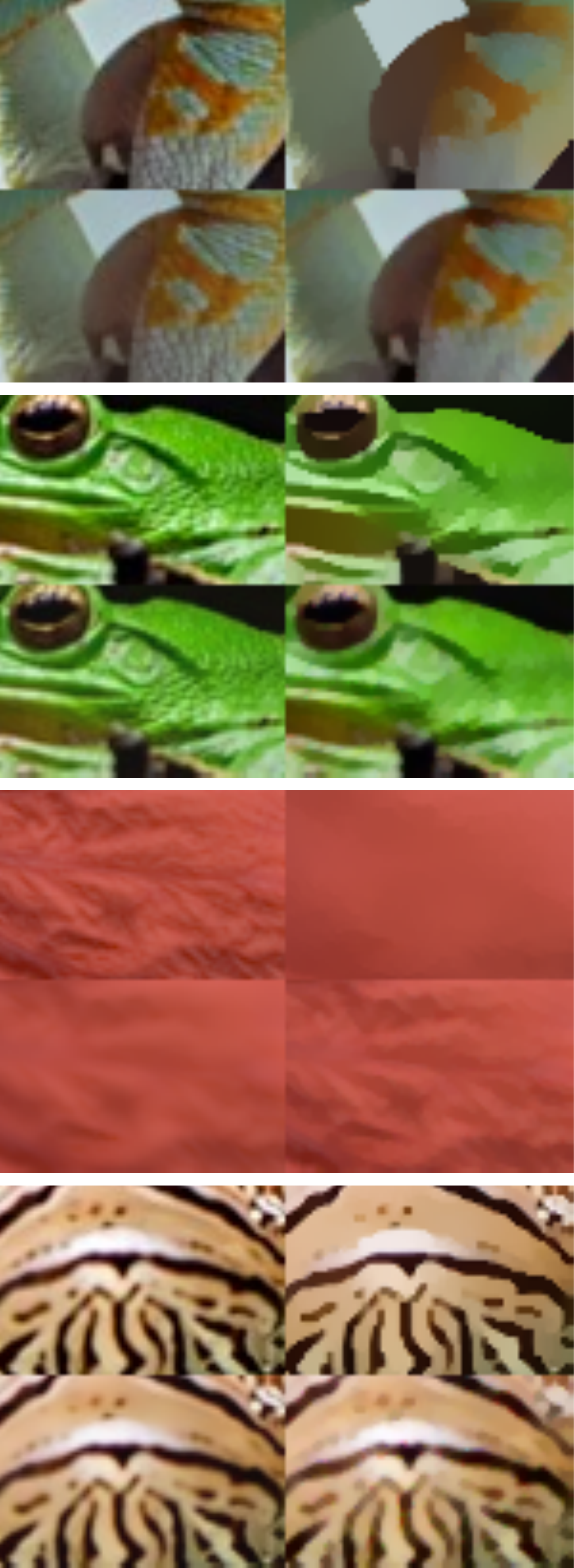}
	\caption{Comparison with other edge preserving filters. From left to right: original, Domain Transform~\cite{DomainTransform}, Guided Filter~\cite{he2010guided}, our method with 10 iterations, and close-ups of the results (in the same order, row-first from top-left to bottom-right). All method parameters were set as in their original papers.
    In image~1 at the foot region, DT smooths all details and generates block artifacts, GF loses the color contrast, while our proposed method successfully removes the texture detail preserving major structures. 
    Similarly, our method successfully removes the details in image~2 without generating block artifacts nor losing color contrast; preserves the color contrast in image~3 the best; and removes the dots in image~4 while keeping the stripes. }
	\label{fig:compare}
\end{figure*} 

\subsection{Compared with Other Edge Preserving Filters}
One iteration of the mean curvature flow with our discrete computation scheme can be considered as a filter. 
Therefore, our filter is comparable with other edge preserving filters, such as Domain Transform (DT)~\cite{DomainTransform} and Guided Filter (GF)~\cite{he2010guided}. 
We tested these three filters on four images as shown in Fig.~\ref{fig:compare}.
As seen with these results, DT tends to generate block artifacts while GF may lose the color contrast.
Our method successfully removes the details while preserving the major structures, without any such side-effects.

\section{Conclusion}
We presented weighted mean curvature and showed its advantageous properties such as scale- and sampling-invariance. 
It can be used as an approximation of gradient for area regularization, which facilitates optimization in image processing applications. 
We discretized the computation with 8 kernels with directional normals, which avoids numerical issues and leads to a very efficient computation scheme.
Our experiments confirm the benefits from the proposed WMC and its discrete computation scheme, including scale- and contrast-robustness, fast computation, and edge preservation.
We compared mean curvature flow computed by a classical method and our new scheme, demonstrating better edge preservation with the latter. 
We also presented our scheme as a neural network layer with its performance on natural images and potential to learn these filters. 
In comparison to other edge preserving filters, our method can produce higher quality results.  

As being independent from any particular imaging model, WMC and area regularization can both be applied on a large range of image processing problems, including smoothing, denoising, super resolution, deconvolution, and image reconstruction. 
In most of these problems, minimizing the area regularization becomes the bottleneck, where our efficient WMC approximation can significantly accelerate this procedure. 
Our computation scheme can be further extended for higher dimensional data such as video and 3D images. 
Thanks to the high performance, our method can be adopted for real time image processing tasks or used on embedded devices such as FPGA, with potential applications in mobile applications such as on smart phones, medical devices, and microscopes.

\section*{Acknowledgments}
Funding was provided by the Swiss National Science Foundation (SNSF) and the Titan X GPU was sponsored by the NVIDIA Corporation.
The authors wish to thank Dr.\ Valeriy Vishnevskiy for his feedback on the convergence of the fast-TV implementation.

\appendix
\section{Matlab Code for computing WMC}
\vspace{-3ex}
\footnotesize
\begin{lstlisting}
function Hw=WeightedMeanCurvature(u)
%compute weighted mean curvature
k1=[1,1,0;  2, -6,0;  1,1,0]/6;
k2=[2,4,1;  4,-12,0;  1,0,0]/12;
dist = zeros([size(u),8],'single');
dist(:,:,1) = conv2(u,k1,'same');
dist(:,:,2) = conv2(u,fliplr(k1),'same');
dist(:,:,3) = conv2(u,k1','same');
dist(:,:,4) = conv2(u,flipud(k1'),'same');
dist(:,:,5) = conv2(u,k2,'same');
dist(:,:,6) = conv2(u,fliplr(k2),'same');
dist(:,:,7) = conv2(u,flipud(k2),'same');
dist(:,:,8) = conv2(u,rot90(k2,2),'same');
%% find minimum signed distance
[~,ind] = min(abs(dist),[],3);
%turn sub to index (sub2ind but faster)
N=size(u,1)*size(u,2); 
offset=int32(reshape(-(N-1):0,size(u,1),size(u,2)));
index=offset+int32(ind*N);
Hw = dist(index); 
\end{lstlisting}
\vspace{2ex}

\bibliographystyle{IEEEtranS}
\bibliography{WMC}

\begin{thebibliography}{10}
\providecommand{\url}[1]{#1}
\csname url@samestyle\endcsname
\providecommand{\newblock}{\relax}
\providecommand{\bibinfo}[2]{#2}
\providecommand{\BIBentrySTDinterwordspacing}{\spaceskip=0pt\relax}
\providecommand{\BIBentryALTinterwordstretchfactor}{4}
\providecommand{\BIBentryALTinterwordspacing}{\spaceskip=\fontdimen2\font plus
\BIBentryALTinterwordstretchfactor\fontdimen3\font minus
  \fontdimen4\font\relax}
\providecommand{\BIBforeignlanguage}[2]{{%
\expandafter\ifx\csname l@#1\endcsname\relax
\typeout{** WARNING: IEEEtranS.bst: No hyphenation pattern has been}%
\typeout{** loaded for the language `#1'. Using the pattern for}%
\typeout{** the default language instead.}%
\else
\language=\csname l@#1\endcsname
\fi
#2}}
\providecommand{\BIBdecl}{\relax}
\BIBdecl

\bibitem{Arbelaez2011}
P.~Arbelaez, M.~Maire, C.~Fowlkes, and J.~Malik, ``Contour detection and
  hierarchical image segmentation,'' \emph{IEEE Transactions on Pattern
  Analysis and Machine Intelligence}, vol.~33, no.~5, pp. 898--916, May 2011.

\bibitem{Carlos:2010}
C.~Brito-Loeza and K.~Chen, ``Multigrid algorithm for high order denoising,''
  \emph{SIAM Journal on Imaging Sciences}, vol.~3, no.~3, pp. 363--389, 2010.

\bibitem{chambolle:2011}
A.~Chambolle and T.~Pock, ``A first-order primal-dual algorithm for convex
  problems with applications to imaging,'' \emph{Journal of Mathematical
  Imaging and Vision}, vol.~40, pp. 120--145, 2011.

\bibitem{chen1991}
Y.~G. Chen, Y.~Giga, and S.~Goto, ``Uniqueness and existence of viscosity
  solutions of generalized mean curvature flow equations,'' \emph{J.
  Differential Geom.}, vol.~33, no.~3, pp. 749--786, 1991.

\bibitem{LLS}
A.~Ciomaga, P.~Monasse, and J.~M. Morel, ``Level lines shortening yields an
  image curvature microscope,'' in \emph{2010 IEEE International Conference on
  Image Processing}, Sept 2010, pp. 4129--4132.

\bibitem{Crane:2013}
K.~Crane, U.~Pinkall, and P.~Schr\"{o}der, ``Robust fairing via conformal
  curvature flow,'' \emph{ACM Trans. Graph.}, vol.~32, 2013.

\bibitem{el20172d}
S.~D. El~Hadji, R.~Alexandre, and A.-O. Boudraa, ``2d curvature-based analysis
  of intrinsic mode functions,'' \emph{IEEE Signal Processing Letters},
  vol.~PP, no.~99, pp. 1--1, August 2017.

\bibitem{DomainTransform}
E.~S.~L. Gastal and M.~M. Oliveira, ``Domain transform for edge-aware image and
  video processing,'' \emph{ACM TOG}, vol.~30, no.~4, pp. 69:1--69:12, 2011,
  proceedings of SIGGRAPH 2011.

\bibitem{goldstein2009geometric}
T.~Goldstein, X.~Bresson, and S.~Osher, ``Geometric applications of the split
  {B}regman method: Segmentation and surface reconstruction,'' \emph{J. Sci.
  Comput.}, vol.~45, no.~1, pp. 272--293, 2009.

\bibitem{gong:gdp}
Y.~Gong and I.~Sbalzarini, ``A natural-scene gradient distribution prior and
  its application in light-microscopy image processing,'' \emph{IEEE J Selected
  Topics in Signal Processing}, vol.~10, no.~1, pp. 99--114, 2016.

\bibitem{gong:phd}
Y.~Gong, ``Spectrally regularized surfaces,'' Ph.D. dissertation, ETH Zurich,
  Nr. 22616, 2015, http://dx.doi.org/10.3929/ethz-a-010438292.

\bibitem{gong:Bernstein}
------, ``Bernstein filter: A new solver for mean curvature regularized
  models,'' in \emph{2016 IEEE International Conference on Acoustics, Speech
  and Signal Processing (ICASSP)}, 2016, pp. 1701--1705.

\bibitem{gong2013a}
Y.~Gong and I.~F. Sbalzarini, ``Local weighted {G}aussian curvature for image
  processing,'' \emph{Intl. Conf. Image Proc. (ICIP)}, pp. 534--538, 2013.

\bibitem{gong:cf}
------, ``Curvature filters efficiently reduce certain variational energies,''
  \emph{IEEE Transactions on Image Processing}, vol.~26, no.~4, pp. 1786--1798,
  2017.

\bibitem{gong2009symmetry}
Y.~Gong, Q.~Wang, C.~Yang, Y.~Gao, and C.~Li, ``Symmetry detection for
  multi-object using local polar coordinate,'' \emph{Lecture Notes in Computer
  Science}, vol. 5702, p. 277, 2009.

\bibitem{Graber2015}
G.~Graber, J.~Balzer, S.~Soatto, and T.~Pock, ``Efficient minimal-surface
  regularization of perspective depth maps in variational stereo,'' in
  \emph{2015 IEEE Conference on Computer Vision and Pattern Recognition
  (CVPR)}, June 2015, pp. 511--520.

\bibitem{he2010guided}
K.~He, J.~Sun, and X.~Tang, ``Guided image filtering,'' \emph{ECCV 2010}, pp.
  1--14, 2010.

\bibitem{huisken2001}
\BIBentryALTinterwordspacing
G.~Huisken and T.~Ilmanen, ``The inverse mean curvature flow and the riemannian
  penrose inequality,'' \emph{J. Differential Geom.}, vol.~59, no.~3, pp.
  353--437, 11 2001. [Online]. Available:
  \url{http://projecteuclid.org/euclid.jdg/1090349447}
\BIBentrySTDinterwordspacing

\bibitem{jia2010fast}
R.-Q. Jia and H.~Zhao, ``A fast algorithm for the total variation model of
  image denoising,'' \emph{Advances in Computational Mathematics}, vol.~33,
  no.~2, pp. 231--241, 2010.

\bibitem{jing-yi:2013}
C.~Jing-yi, ``The willmore functional of surfaces,'' \emph{Appl. Math.-J. Chin.
  Univ. Ser. B}, vol.~28, no.~4, pp. 485--493, December 2013.

\bibitem{isoLaplace:1999}
B.~Kamgar-Parsi, B.~Kamgar-Parsi, and A.~Rosenfeld, ``Optimally isotropic
  laplacian operator,'' \emph{IEEE Transactions on Image Processing}, vol.~8,
  no.~10, pp. 1467--1472, Oct 1999.

\bibitem{OsherTV2005}
S.~Osher, M.~Burger, D.~Goldfarb, J.~Xu, and W.~Yin, ``An iterative
  regularization method for total variation-based image restoration,''
  \emph{Multiscale Model. Simul.}, vol.~4, no.~2, pp. 460--489, 2005.

\bibitem{patra2006stencils}
M.~Patra and M.~Karttunen, ``Stencils with isotropic discretization error for
  differential operators,'' \emph{Numerical Methods for Partial Differential
  Equations}, vol.~22, no.~4, pp. 936--953, 2006.

\bibitem{RenQL15}
F.~Ren, T.~Qiu, and H.~Liu, ``Mean curvature regularization-based {P}oisson
  image restoration,'' \emph{J. Electronic Imaging}, vol.~24, no.~3, p. 033025,
  2015.

\bibitem{TV1992}
L.~I. Rudin, S.~Osher, and E.~Fatemi, ``Nonlinear total variation based noise
  removal algorithms,'' \emph{Physica D}, vol.~60, no.~1, pp. 259--268, 1992.

\bibitem{shen2003}
J.~Shen, S.~H. Kang, and T.~F. Chan, ``{E}uler's elastica and curvature-based
  inpainting,'' \emph{SIAM J. Appl. Math.}, vol.~63, no.~2, pp. 564--592, 2003.

\bibitem{velimirovic:2010}
L.~S. Velimirovic, M.~S. Ciric, and M.~D. Cvetkovic, ``Change of the willmore
  energy under infinitesimal bending of membranes,'' \emph{Comput. Math.
  Appl.}, vol.~59, no.~12, pp. 3679--3686, June 2010.

\bibitem{yang:2014}
F.~Yang, K.~Chen, B.~Yu, and D.~Fang, ``A relaxed fixed point method for a mean
  curvature-based denoising model,'' \emph{Optim. Method Softw.}, vol.~29,
  no.~2, pp. 274--285, 2014.

\bibitem{MC:1998}
A.~Yezzi, ``Modified curvature motion for image smoothing and enhancement,''
  \emph{IEEE Transactions on Image Processing}, vol.~7, no.~3, pp. 345--352,
  Mar 1998.

\bibitem{zhang_chen_chen_yu_2017}
J.~Zhang, K.~Chen, F.~Chen, and B.~Yu, ``An efficient numerical method for mean
  curvature-based image registration model,'' \emph{East Asian Journal on
  Applied Mathematics}, vol.~7, no.~1, pp. 125--142, 2017.

\bibitem{Zhu2007}
H.~Zhu, H.~Shu, J.~Zhou, X.~Bao, and L.~Luo, ``Bayesian algorithms for {PET}
  image reconstruction with mean curvature and {G}auss curvature diffusion
  regularizations,'' \emph{Computers in Biology and Medicine}, vol.~37, no.~6,
  pp. 793--804, 2007.

\bibitem{meanZhu}
W.~Zhu and T.~Chan, ``Image denoising using mean curvature of image surface,''
  \emph{SIAM Journal on Imaging Sciences}, vol.~5, no.~1, pp. 1--32, 2012.

\bibitem{zhu:2013}
W.~Zhu, X.-C. Tai, and T.~Chan, ``Augmented {L}agrangian method for a mean
  curvature based image denoising model,'' \emph{Inverse Probl. Imaging},
  vol.~7, no.~4, pp. 1409--1432, 2013.

\bibitem{zimmerberg:2006}
J.~Zimmerberg and M.~M. Kozlov, ``How proteins produce cellular membrane
  curvature,'' \emph{Nat. Rev. Mol. Cell Biol.}, vol.~7, no.~1, pp. 9--19,
  2006.

\end{thebibliography}

\end{document}